\documentclass{article}

\PassOptionsToPackage{numbers, compress}{natbib}
\usepackage[final]{neurips_2022}

\usepackage{captdef}
\usepackage{microtype}
\usepackage{graphicx}
\usepackage{subfigure}
\usepackage{booktabs}
\usepackage{multirow}
\usepackage{booktabs} 
\usepackage{bbm}
\usepackage[table,xcdraw]{xcolor}
\usepackage{hyperref}
\usepackage{amsmath}
\usepackage{amssymb}
\usepackage{mathtools}
\usepackage{amsthm}
\usepackage{multirow}
\usepackage[capitalize,noabbrev]{cleveref}
\usepackage{xcolor,pifont}
\usepackage[textsize=tiny]{todonotes}
\usepackage{wrapfig}

\usepackage{url}            
\usepackage{booktabs}       
\usepackage{amsfonts}       
\usepackage{nicefrac}       
\usepackage{microtype}      
\usepackage{xcolor}         
\usepackage[utf8]{inputenc} 
\usepackage[T1]{fontenc}    
\usepackage{hyperref}       
\usepackage{url}            
\usepackage{booktabs}       
\usepackage{amsfonts}       
\usepackage{nicefrac}       
\usepackage{microtype}      
\usepackage{multirow}
\usepackage{graphicx}
\usepackage{epsfig}
\usepackage{amsmath}
\usepackage{amssymb}
\usepackage{subfigure}
\usepackage{marvosym}
\usepackage{color}
\usepackage{xcolor}         
\usepackage{threeparttable}
\usepackage{algorithm}
\usepackage{algpseudocode}
\usepackage{setspace}

\title{Debiased Self-Training for Semi-Supervised Learning}

\author{%
  \textnormal{Baixu Chen\thanks{Equal contribution.}, Junguang Jiang\footnotemark[1], Ximei Wang, Pengfei Wan$^{\S}$, Jianmin Wang, Mingsheng Long\textsuperscript{\Letter}} \\
  School of Software, BNRist, Tsinghua University, China\\
  \textsuperscript{\S}Y-tech, Kuaishou Technology\\
  {\small \texttt{\{chenbx18,jjg20\}@mails.tsinghua.edu.cn, \{jimwang,mingsheng\}@tsinghua.edu.cn}} \\
}

\begin{document}

\maketitle

\begin{abstract}
  Deep neural networks achieve remarkable performances on a wide range of tasks with the aid of large-scale labeled datasets. Yet these datasets are time-consuming and labor-exhaustive to obtain on realistic tasks. To mitigate the requirement for labeled data, self-training is widely used in semi-supervised learning by iteratively assigning pseudo labels to unlabeled samples. Despite its popularity, self-training is well-believed to be unreliable and often leads to training instability. Our experimental studies further reveal that the bias in semi-supervised learning arises from both the problem itself and the inappropriate training with potentially incorrect pseudo labels, which accumulates the error in the iterative self-training process. To reduce the above bias, we propose \textit{Debiased Self-Training (DST)}. First, the generation and utilization of pseudo labels are decoupled by two parameter-independent classifier heads to avoid direct error accumulation. Second, we estimate the worst case of self-training bias, where the pseudo labeling function is accurate on labeled samples, yet makes as many mistakes as possible on unlabeled samples. We then adversarially optimize the representations to improve the quality of pseudo labels by avoiding the worst case. Extensive experiments justify that \textit{DST} achieves an average improvement of $6.3\%$ against state-of-the-art methods on standard semi-supervised learning benchmark datasets and $18.9\%$ against FixMatch on $13$ diverse tasks. Furthermore, DST can be seamlessly adapted to other self-training methods and help stabilize their training and balance performance across classes in both cases of training from scratch and finetuning from pre-trained models.
\end{abstract}

\section{Introduction}
Deep learning has achieved great success in many machine learning problems in the past decades, especially where large-scale labeled datasets are present. In real-world applications, however, manually labeling sufficient data is time-consuming and labor-exhaustive.
To reduce the requirement for labeled data, semi-supervised learning (SSL)  improves the data efficiency of deep models by learning from a few labeled samples and a large number of unlabeled samples~\citep{entropy_minimization, pseudo_label, Mean_Teacher, Simclrv2}.
Among them, self-training is an effective approach to deal with the lack of labeled data.
Typical self-training methods \citep{pseudo_label, FixMatch} assign pseudo labels to unlabeled samples with the model's predictions and then iteratively train the model with these pseudo labeled samples as if they were labeled examples.

Although self-training has achieved great advances in benchmark datasets, they still exhibit large training instability and extreme performance imbalance across classes.
For instance, the accuracy of FixMatch \citep{FixMatch}, one of the state-of-the-art self-training methods, fluctuates greatly when trained \emph{from scratch} (see Figure \ref{fig:scratch_loss_and_acc}). Though its performance will gradually recover after a sudden sharp drop, this is still not expected, since \textit{pre-trained} models are more often adopted~\citep{cite:NAACL19BERT,  Simclrv2, jiang2022transferability} are improve data efficiency, and the performance of pre-trained models is difficult to recover after a drastic decline due to catastrophic forgetting \citep{catastrophic_forgetting}.
Besides, although FixMatch improves the average accuracy, it also leads to the \textit{Matthew effect}, \textit{i.e.},
the accuracy of well-behaved categories is further increased while that of poorly-behaved ones is decreased to nearly zero (see Figure \ref{fig:per_cls}). This is also not expected, since most machine learning models prefer performance balance across categories, even when the class imbalance exists in the training data \citep{place}. 
The above findings are caused by the \emph{bias} between the pseudo labeling function with the unknown target labeling function.
Training with biased and unreliable pseudo labels has the chance to accumulate errors and ultimately lead to performance fluctuations.
And for those poorly-behaved categories, the bias of the pseudo labels gets worse and will be further enhanced as self-training progresses, ultimately leading to the Matthew effect.


We delved into the bias issues arising from the self-training process and found that they can be briefly grouped into two kinds: 
(1) \textit{Data bias}, the bias inherent in the SSL tasks;
(2) \textit{Training bias}, the bias increment brought by self-training with incorrect pseudo labels.
In this regard, we present \textit{Debiased Self-Training (DST)}, a novel approach to decrease the undesirable bias in self-training.
Specifically, to reduce the \textit{training bias}, the classifier head is only trained with clean labeled samples and no longer trained with unreliable pseudo-labeled samples.
In other words, the generation and utilization of pseudo labels are decoupled to mitigate bias accumulation and boost the model's tolerance to biased pseudo labels.
Further, to decrease the \textit{data bias} which cannot be calculated directly, we turn to estimate the worst case of training bias that implicitly reflects the data bias. Then we optimize the representations to decrease the worst-case bias and thereby improve the quality of pseudo labels.

The contributions of this work are summarized as follows:
(1) We systematically identify the problem and analyze the causes of self-training bias in semi-supervised learning.
(2) We propose \textit{DST}, a novel approach to mitigate the self-training bias and boost the stability and performance balance across classes, which can be used as a universal add-on for mainstream self-training methods.
(3) We conduct extensive experiments and validate that \textit{DST} achieves an average boost of $6.3\%$ against state-of-the-art methods on standard datasets and $18.9\%$ against FixMatch on $13$ diverse tasks.



\section{Related Work}
\subsection{Self-training for  semi-supervised learning}
Self-training \cite{self_training_acl, self_training_wacv, entropy_minimization, pseudo_label} is a widely-used approach to utilize unlabeled data. Pseudo Label \cite{pseudo_label}, one popular self-training method, iteratively generates pseudo labels and utilizes them with the same model. However, this paradigm suffers from the problem of confirmation bias \cite{pseudoLabel2019}, where the learner struggles to correct its own mistakes when learning from inaccurate pseudo labels.
The bias issue is also mentioned in DebiasMatch \cite{DebiasMatch} where they define the bias as the quantity imbalance for each category.  Note that the bias in our paper refers to 
the deviation between the pseudo labeling function and the ground truth labeling function, which is a more essential problem existing in most self-training methods.
Recent works mainly tackle this bias issue from the following two aspects.



\textbf{Generate higher-quality pseudo labels.}
MixMatch \cite{MixMatch} averages predictions from multiple augmentations as pseudo labels. ReMixMatch \cite{ReMixMatch}, UDA \cite{UDA}, and FixMatch \cite{FixMatch} adopt  confidence thresholds to generate pseudo labels on weakly augmented samples and utilize these pseudo-labels as annotations for strongly augmented samples. 
Dash \cite{Dash} and FlexMatch \cite{FlexMatch} dynamically adjust the thresholds in a curriculum learning manner. Label Propagation methods \cite{transductive, label_propagation} assign pseudo labels with the density of the local neighborhood. DASO \cite{DASO} blends the confidence-based pseudo labels and density-based pseudo labels differently for each class. Meta Pseudo Labels \cite{meta_pseudo_labels} proposes to generate pseudo labels with a meta learner. Different from the above methods that manually design specific criteria to improve the quality of pseudo labels, 
we estimate the \textit{worst case} of self-training bias and adversarially optimize the representations to improve the quality of pseudo labels automatically.


\textbf{Improve tolerance to inaccurate pseudo labels.}
To mitigate the confirmation bias, existing methods maintain a mismatch between the generation and utilization of pseudo labels. Temporal Ensembling \cite{Temporal_Ensembling} and Mean Teacher \cite{Mean_Teacher} generate pseudo labels from the average of previous predictions or an exponential moving average of the model, respectively. Noisy Student \cite{Noisy_Student} assigns pseudo labels by a fixed teacher from the previous round. Co-training \cite{CoTraining}, MMT \cite{MMT}, DivideMix \cite{DivideMix} and Multi-head Tri-training \cite{MTTriTraining} introduce multiple models or classifier heads and learn in an online mutual-teaching manner. 
In these methods, each classifier head is still trained with potentially incorrect pseudo labels generated by other heads. In contrast, in our method, \textit{the classifier head that generates pseudo labels is never trained with pseudo labels}, leading to better tolerance to inaccurate pseudo labels (Table \ref{table:ablation}).

\subsection{Self-supervised learning for  semi-supervised learning}

Self-supervised methods \citep{cite:NAACL19BERT, cite:CVPR20MoCo} are also used on unlabeled data to improve the model with few labeled samples, either in the pre-training stage \citep{Simclrv2, PAWS} or in the downstream tasks \citep{Self-Tuning, CoMatch}. However, the training of self-supervision usually relies on big data and heavy computation, which is not feasible in most applications. Besides, although these methods avoid the use of unreliable pseudo labels, it is difficult for them to learn task-specific information from unlabeled data for better performance.

\subsection{Adversarial training for semi-supervised learning}

Some works introduce adversarial training \cite{GAN} into semi-supervised learning. 
A line of works \cite{GANSSL, ImproveGANs, GoodSSLBadGAN, ALI} exploit fake samples from the generator by labeling them with a new ``generated'' class and forcing the discriminator to output class labels. Another line of works use adversarial training to construct adversarial samples \cite{adversarial_samples}, e.g., VAT \cite{VAT} injects additive noise into input, VAdD \cite{VAdD} introduces adversarial Dropout \cite{Dropout} layers and RAT \cite{RAT} expands the noise in VAT into a set of input transformations. These methods aim to impose a local smoothness on the model and do not involve training with pseudo labels. In contrast, in our method, the goal of the adversarial training is to estimate the worst case of pseudo labeling and then avoid such cases (Section \ref{sec:worst_case}).

\begin{figure*}[!b]
    \centering
    \includegraphics[width=1\textwidth]{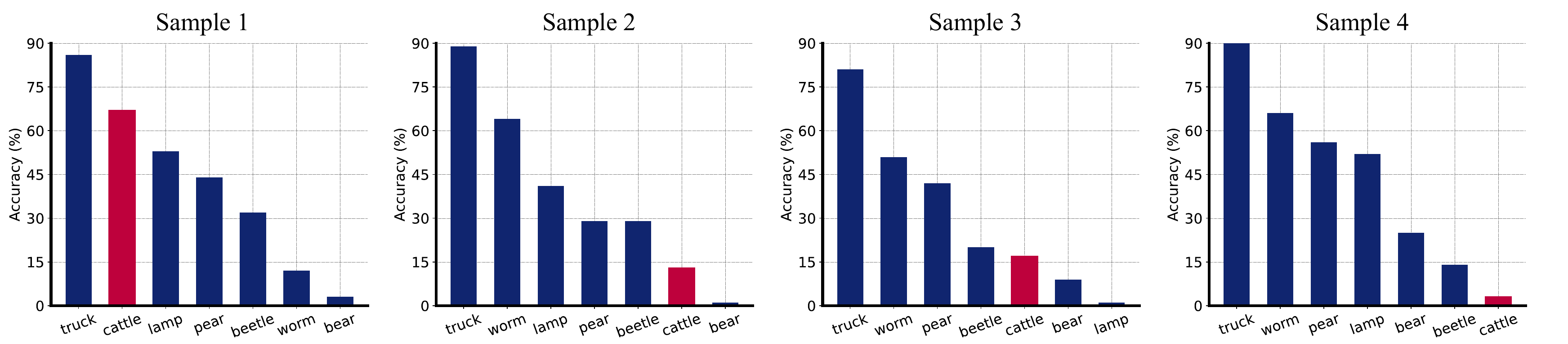}
    \vspace{-10pt}
    \caption{Effect of \textit{data sampling}. Top-1 accuracy of $7$ randomly selected categories when trained with different labeled data sampled from \emph{CIFAR-100}. 
    The same category (such as \textcolor{red}{cattle}) may have completely different accuracy in different samples.
    Following FixMatch \cite{FixMatch}, $4$ labeled data are sampled for each category by default in our analysis. 
    }
    \label{fig:data_bias}
\end{figure*}

\begin{figure*}[!b]
\centering
\begin{minipage}[c]{0.47\textwidth}
\centering
    \includegraphics[width=1\textwidth]{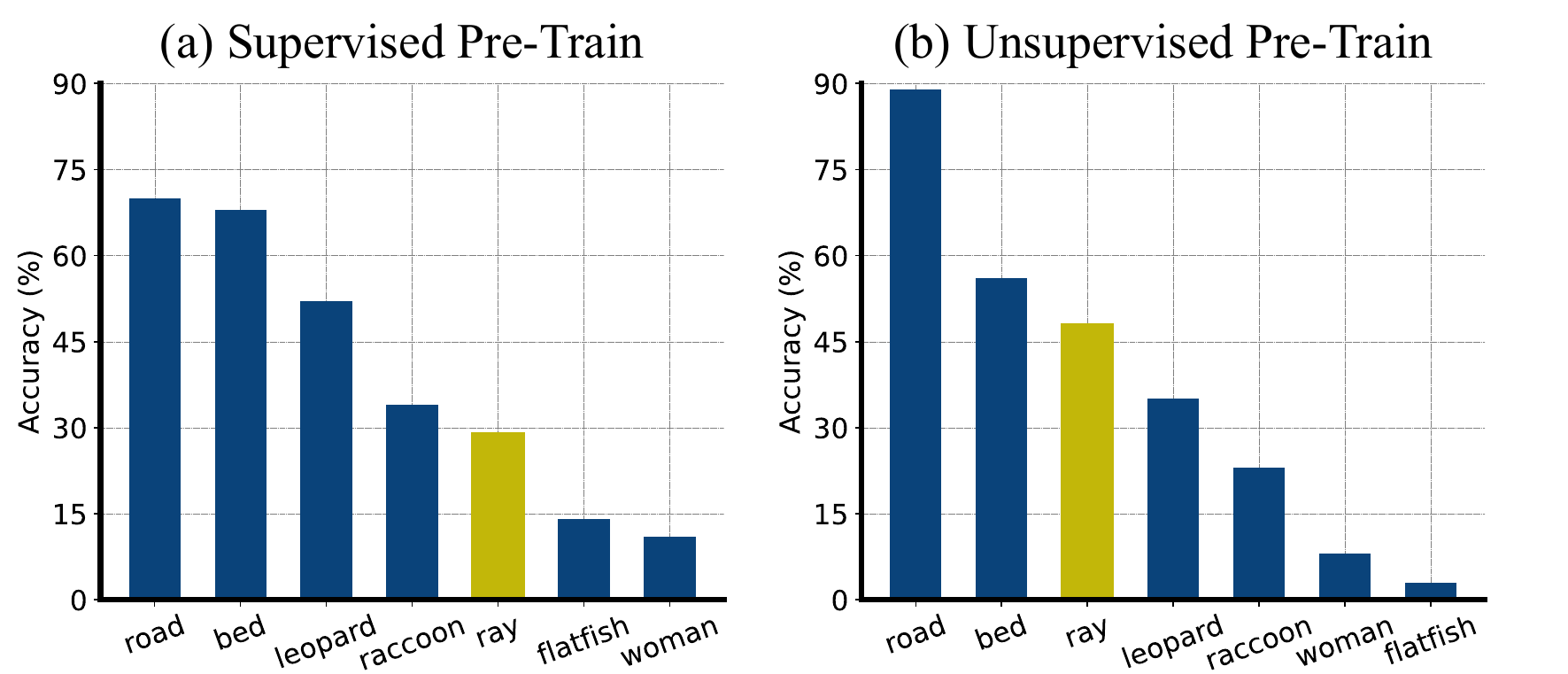}
    \vspace{-10pt}
    \caption{Effect of \textit{pre-trained representations}.
    Accuracy of $7$ randomly selected categories with different pre-trained models on \emph{CIFAR-100}.
    Different pre-trained models show different category preferences. }
    \label{fig:model_bias}
\end{minipage}
\hspace{14pt}
\begin{minipage}[c]{0.47\textwidth}
    \includegraphics[width=1\textwidth]{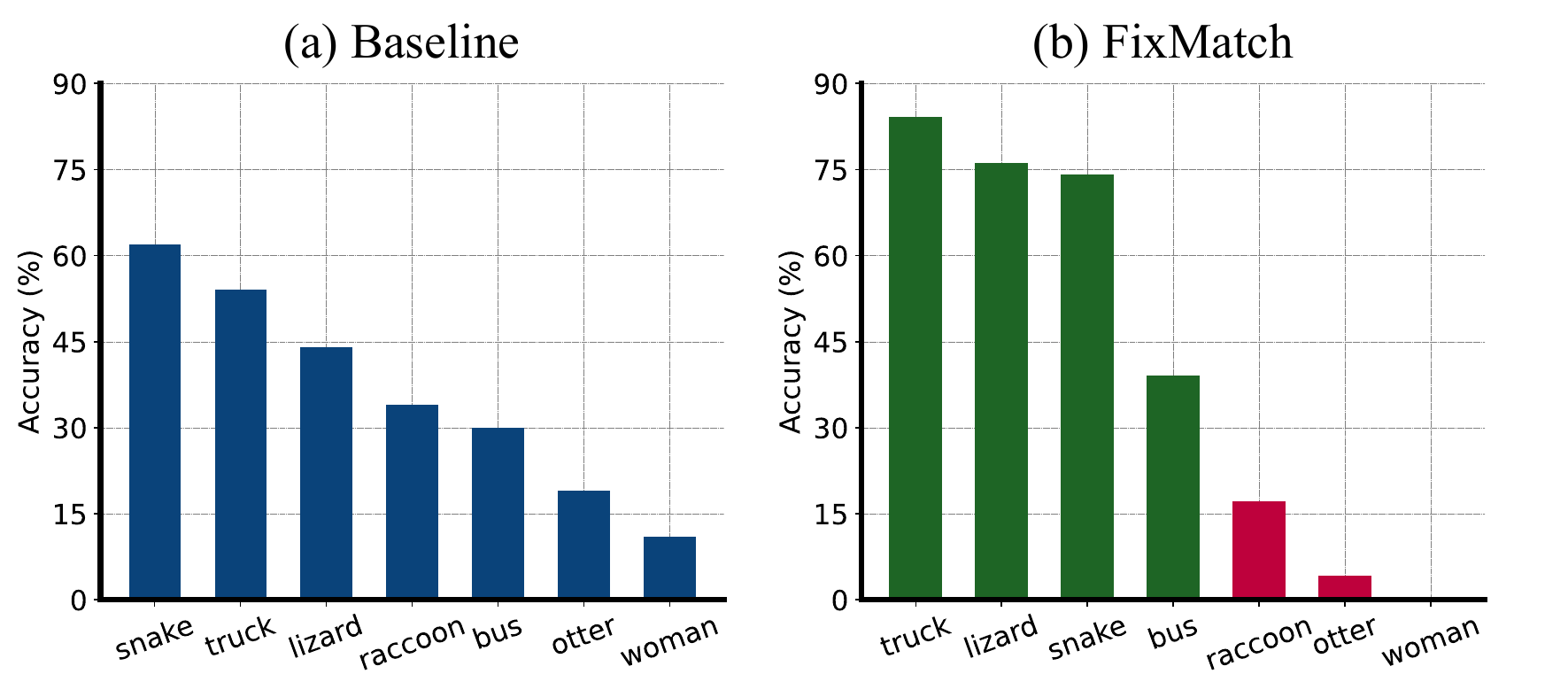}
    \vspace{-10pt}
    \caption{Effect of \textit{self-training algorithm}. Accuracy of $7$ randomly selected categories with different training methods on \emph{CIFAR-100}. FixMatch largely increases the bias of poorly-behaved categories (Matthew effect). }
    \label{fig:training_bias}
\end{minipage}
\vspace{-5pt}
\end{figure*}

\section{Analysis of Bias in Self-Training}
\label{sec:analysis}
In this section, we provide some analysis of where the bias in self-training comes from.
Let $\mathcal{P}$ denote a distribution over input space $\mathcal{X}$.
For classification with $K$ classes, 
let $P^k$ denote the class-conditional distribution of $\mathbf{x}$ conditioned on ground truth $f^*(\mathbf{x})=k$.  
Assume that pseudolabeler $f_{\text{pl}}$ is obtained via training a classifier on $n$ labeled samples $\widehat{P}_n$.
Let $\mathcal{M}(f_{\text{pl}})\triangleq \{\mathbf{x}: f_{\text{pl}}(\mathbf{x}) \neq f^*(\mathbf{x})\}$ denote the mistaken pseudolabeled samples.
The bias in the self-training refers to \textit{the deviation between the learned decision hyperplanes and the true decision hyperplanes}, which can be measured by 
the fraction of incorrectly pseudolabeled samples in any classes $\mathcal{B}(f_{\text{pl}}) = \{ P^k(\mathcal{M}(f_{\text{pl}})) \}_{k=1}^{K}$ \citep{self_training_theory}.
By analyzing self-training bias under different training conditions, we have several nontrivial findings.

\textit{The sampling of labeled data will largely influence the self-training bias.}
As shown in Figure \ref{fig:data_bias}, when the data sampling is different, the accuracy of the same category may vary dramatically. 
The reason is that the distances between different data points and the true decision hyperplanes are not the same, with some supporting data points closer and others far away. 
When there are few labeled data, there may be a big difference in the distances between supporting data of each category and the true decision hyperplanes, hence the learned decision hyperplanes will be biased towards some categories.

\textit{The pre-trained representations also affect the self-training bias.} Figure \ref{fig:model_bias} shows that different pre-trained representations lead to different category bias, even if the pre-trained dataset and the downstream labeled dataset are both identical.
One possible reason is that the representations learned by different pre-trained models focus on different aspects of the data  \citep{cite:ICLR21InstanceTransfer}. Therefore, the same data could also have different distances to the decision hyperplanes in the representation level with different pre-trained models.


\textit{Training with pseudo labels aggressively in turn enlarges the self-training bias on some categories.} Figure \ref{fig:training_bias} shows that  
after training with pseudo labels (\textit{e.g.}, using FixMatch), 
the performance gap for different categories greatly enlarges,
with the accuracy of some categories increasing from $60\%$ to $80\%$ and that of some categories dropping from $15\%$ to $0\%$.
The reason is that for well-behaved categories, the pseudo labels are almost accurate, hence using them for training could further reduce the bias. Yet for many poorly-behaved categories, the pseudo labels are not reliable, and the common self-training mechanism that uses these incorrect pseudo labels to train the model will further increase the bias, and fail to correct it back in the follow-up training. This results in the Matthew effect.

\begin{figure}[htbp]
    \begin{center}
    \centering
    \vspace{-10pt}
    \includegraphics[width=0.7\linewidth]{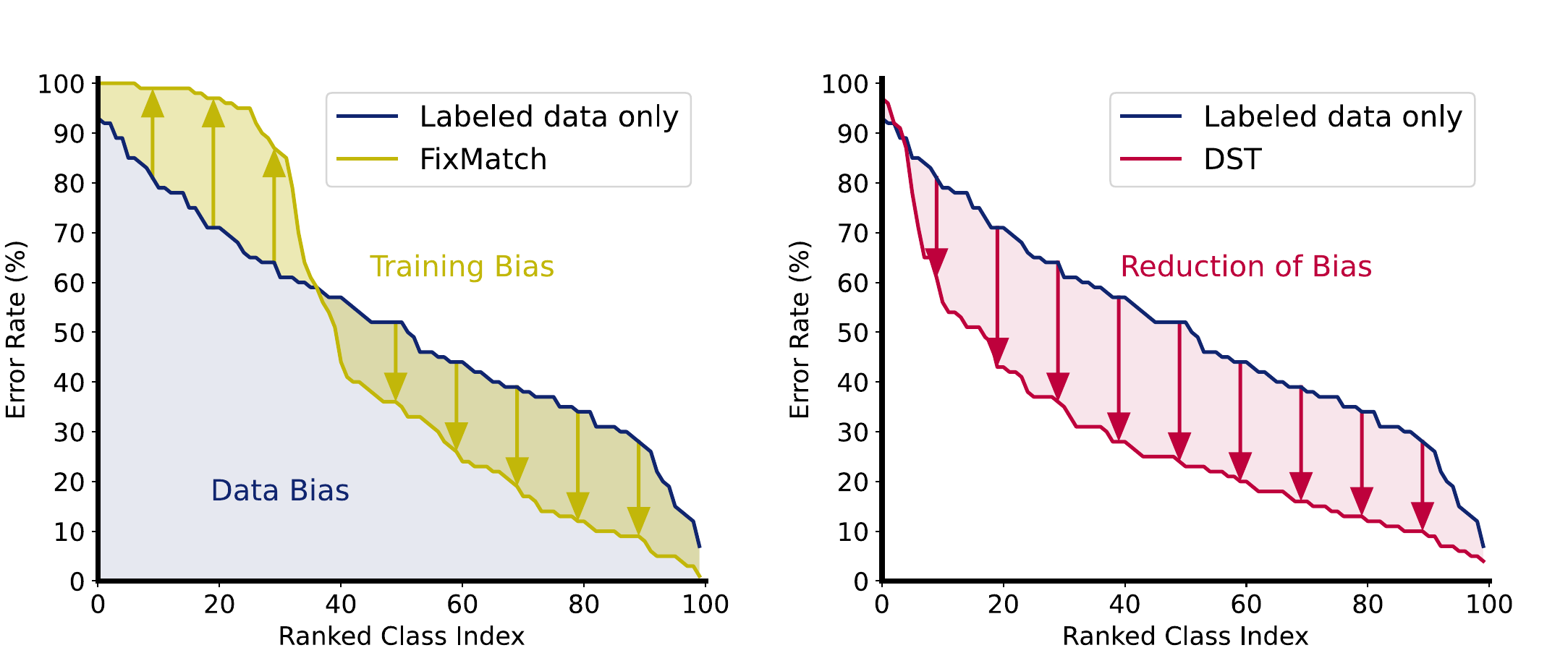}
    \caption{
    Error rate of pseudo labels in any classes on \emph{CIFAR-100} (ResNet50, $4$ labels per category). 
    FixMatch decreases the bias on well-behaved categories while increasing that of poorly-behaved categories. In contrast, DST effectively balances the performance between different categories.
    }
    \label{fig:bias_illustration}
    \label{fig:per_cls}
    \vspace{-5pt}
  \end{center}
\end{figure}

Based on the above observations, we divide the bias caused by self-training into two categories.


\textbf{Data bias}: the bias inherent in semi-supervised learning tasks, such as the bias of sampling and pre-trained representations on unlabeled data.
Formally, data bias
 is defined as  $\mathcal{B}(f_{\text{pl}}(\widehat{P}_n, \psi_0))-\mathcal{B}(f^*)$ (\textcolor{blue}{blue} area in Figure~\ref{fig:bias_illustration}), where the pseudolabeler $f_{\text{pl}}(\widehat{P}_n, \psi_0)$ is obtained from a biased sampling $\widehat{P}_n$ with a biased parameter initialization $\psi_0$.

\textbf{Training bias}: the bias increment brought by some unreasonable training strategies.
Formally, {training bias} is 
    $\mathcal{B}(f_{\text{pl}}(\widehat{P}_n, \psi_0, \mathcal{S}))-\mathcal{B}(f_{\text{pl}}(\widehat{P}_n, \psi_0))$  (\textcolor[rgb]{0.5, 0.5, 0}{yellow} area in Figure~\ref{fig:bias_illustration}) 
    where $f_{\text{pl}}(\widehat{P}_n, \psi_0,\mathcal{S})$ is a pseudolabeler obtained with
    self-training strategy $\mathcal{S}$.

Next we will introduce how to reduce training bias and data bias in self-training (\textcolor{red}{red} area in Figure~\ref{fig:bias_illustration}).

\section{Debiased Self-Training}
\label{sec:debias}
In semi-supervised learning (SSL), we have a labeled dataset $\mathcal{L}=\{(\mathbf{x}_i^l, {y}_i^l \}_{i=1}^{n_l}$ of $n_l$ labeled samples and an unlabeled dataset $\mathcal{U}=\{(\mathbf{x}_j^u) \}_{j=1}^{n_u}$ of $n_u$ unlabeled samples, where the size of the labeled dataset is usually much smaller than that of the unlabeled dataset, \textit{i.e.}, $n_l \ll n_u$. Denote $\psi$ the feature generator, and
$h$ the task-specific head. The standard cross-entropy loss on weakly augmented labeled examples is
\begin{equation}
    L_\mathcal{L}(\psi, h)=\dfrac{1}{n_l} \sum_{i=1}^{n_l}L_{\text{CE}} \big((h \circ \psi \circ \alpha)(\mathbf{x}_i^l), {y}_i^l\big),
\end{equation}
where $\alpha$ is the weak augmentation function.
Since there are few labeled samples,
the feature generator and the task-specific head will easily over-fit,  and typical SSL methods use these pseudo labels on plenty of unlabeled data to decrease the generalization error.  Different SSL methods design different pseudo labeling function $\widehat{f}$ \citep{pseudo_label, Dash, Defense}.
Take FixMatch \citep{FixMatch} for an instance. FixMatch first generates predictions 
$\widehat{\mathbf{p}} = (h\circ \psi \circ \alpha)(\mathbf{x})$
on a weakly augmented version of given unlabeled images, and adopts a confidence threshold $\tau$ to filter out unreliable pseudo labels,
\begin{equation}
    \widehat{f}_{\psi, h}(\mathbf{x}) =
    \begin{cases}
    \arg \max \widehat{\mathbf{p}}, &\max \widehat{\mathbf{p}} \ge \tau, \\
    -1, &\text{otherwise,}
    \end{cases}
\end{equation}
where $\widehat{f}_{\psi, h}$ refers to the pseudo labeling by model $h \circ \psi$,
hyperparameter $\tau$ specifies the threshold above which a pseudo label is retained and $-1$ indicates that this pseudo label is ignored in training.
Then FixMatch utilizes selected pseudo labels to train on strongly augmented unlabeled images,
\begin{equation}
 L_\mathcal{U}(\psi, h, \widehat{f}) = \dfrac{1}{n_u} \sum_{j=1}^{n_u} L_{\text{CE}} \big((h \circ \psi \circ \mathcal{A})(\mathbf{x}_j^u), \widehat{f}(\mathbf{x}_j^u)\big),
\end{equation}
where $\widehat{f}$ is a notation of general pseudo labeling function and $\mathcal{A}$ is the strong augmentation function. As shown in Figure \ref{fig:architectures}(a), the optimization objective for FixMatch is
\begin{equation}
\label{eq:fixmatch_optimization}
    \min_{\psi, h} L_\mathcal{L}(\psi, h) + \lambda L_\mathcal{U}(\psi, h, \widehat{f}_{\psi, h}),
\end{equation}
where $\lambda$ is the trade-off between the loss on labeled data and that on unlabeled data.
FixMatch filters out low-confidence samples during the pseudo labeling process,
yet two issues remain: (1) The pseudo labels are generated and utilized by the same head, which leads to the training bias, \textit{i.e.},  the errors of the model might be amplified as the self-training progresses. 
(2) When trained with extreme few labeled samples,
the problem of unreliable pseudo labeling caused by data bias cannot be ignored anymore even with the confidence threshold mechanism.
To tackle the above issues, we propose two important designs to decrease training bias and data bias in Section \ref{sec:agent_head} and \ref{sec:worst_case} respectively.

\subsection{Generate and utilize pseudo labels independently}
\begin{figure*}[!b]
    \centering
    \includegraphics[width=1\textwidth]{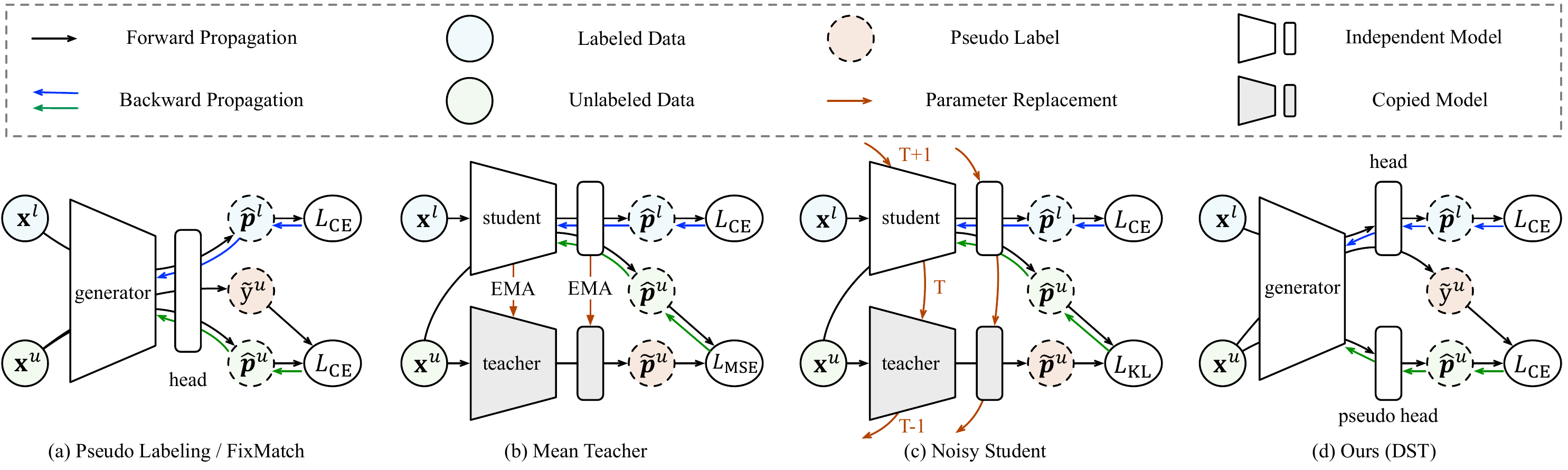}
    \vspace{-10pt}
    \caption{Comparisons on how different self-training methods generate and utilize pseudo labels. 
    \textbf{(a)} Pseudo Labeling and FixMatch generate and utilize pseudo labels on the same model. \textbf{(b)} Mean Teacher generates pseudo labels from the  Exponential Moving Average (EMA) of the current model. \textbf{(c)} Noisy Student generates pseudo labels from the teacher model which is obtained from the previous round of training. \textbf{(d)} DST generates pseudo labels from head $h$ and utilizes pseudo labels on a parameter independent pseudo head $h_{\text{pseudo}}$.}
    \label{fig:architectures}
\end{figure*}

\label{sec:agent_head}
The training bias of FixMatch stems from the way of training on the pseudo labels generated by itself. To alleviate this bias, some methods turn to generate pseudo labels from a better teacher model, such as the moving average of the original model \citep{Mean_Teacher} in Figure \ref{fig:architectures}(b) or the model obtained from the previous round of training \citep{Noisy_Student} in Figure \ref{fig:architectures}(c), and then utilize these pseudo labels to train both the feature generator $\psi$ and the task-specific head $h$.
However, there is still a tight relationship between the teacher model that generates pseudo labels and the student model that utilizes pseudo labels in the above methods, and the decision hyperplanes of the student model $h \circ \psi$ strongly depend on the biased pseudo labeling $\widehat{f}$. As a result, training bias is still large in the self-training process.

To further decrease the training bias when utilizing the pseudo labels, we optimize the task-specific head $h$,
only with the clean labels on $\mathcal{L}$ and without any unreliable pseudo labels from $\mathcal{U}$.  
To prevent the deep models from over-fitting to the few labeled samples, we still use pseudo labels, but only for learning a better representation. As shown in Figure \ref{fig:architectures}(d), we introduce a pseudo head $h_{\text{pseudo}}$, which is connected to the feature generator $\psi$ and only optimized with pseudo labels from $\mathcal{U}$. Then the training objective is
\begin{align}
\label{eq:agent}
\min_{\psi, h, h_{\text{pseudo}}} L_\mathcal{L}(\psi, h) + \lambda L_\mathcal{U}(\psi, h_{\text{pseudo}}, \widehat{f}_{\psi, h}),
\end{align}
where the pseudo labels are generated by head $h$ and utilized by a completely parameter \emph{independent} pseudo head $h_{\text{pseudo}}$.
Although $h$ and $h_{\text{pseudo}}$ are fed with features from the same backbone network, their parameters are independent, thus training the pseudo head $h_{\text{pseudo}}$ with some wrong pseudo labels will not accumulate the bias of head $h$ directly  in the iterative self-training process. 
Note that the pseudo head $h_{\text{pseudo}}$ is only responsible for gradient backpropagation to the feature generator $\psi$ during training and will be discarded during inference, and thus will introduce no inference cost. 



\subsection{Reduce generation of erroneous pseudo labels}
\label{sec:worst_case}
\begin{figure*}[!b]
    \centering
    \includegraphics[width=0.9\textwidth]{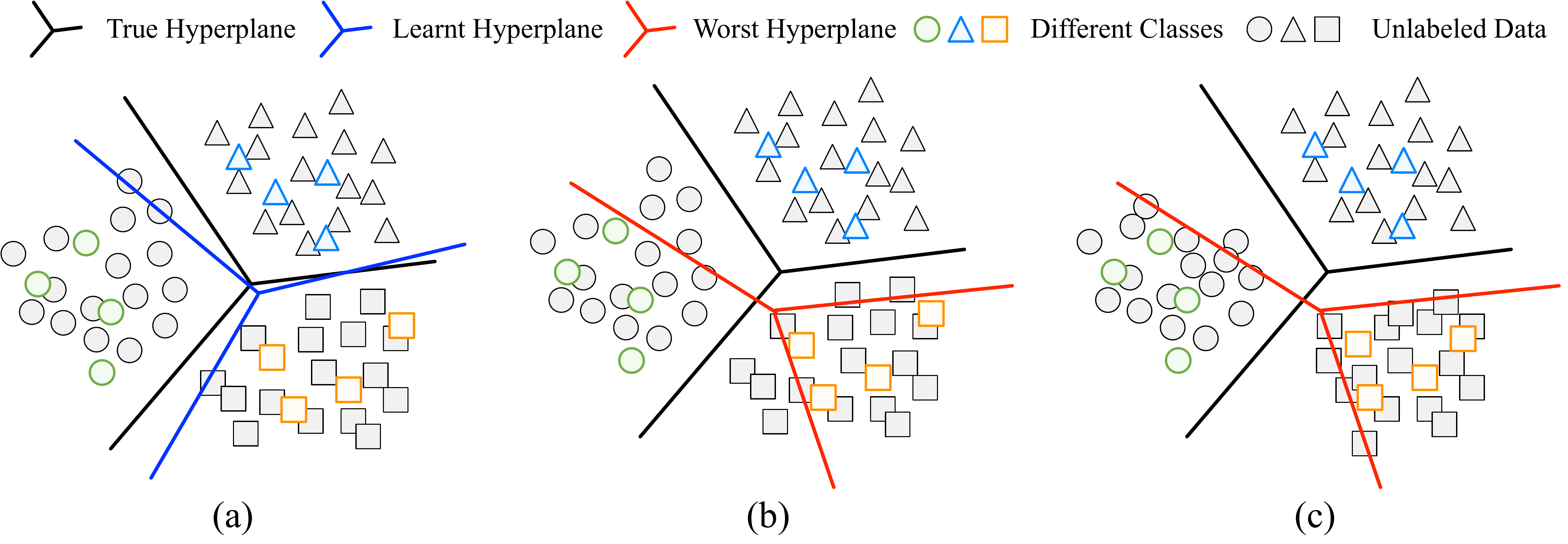}
    \vspace{-5pt}
    \caption{Concept explanations. \textbf{(a)} Shift between {the hyperplanes learned} on limited labeled data and the true hyperplanes. \textbf{(b)} {The worst hyperplanes} are hyperplanes that correctly distinguish labeled samples while making as many mistakes as possible on unlabeled samples. \textbf{(c)} Feature representations are optimized to improve the performance of {the worst hyperplanes}.
}
\label{fig:worst_case}
\end{figure*}

Section \ref{sec:agent_head} presents a solution to reduce the training bias, yet the data bias still exists in the pseudo labeling  $\widehat{f}$.
As shown in Figure \ref{fig:worst_case}(a),  due to the data bias, labeled samples of each class have different distances to the decision hyperplanes in the representation space, which leads to a deviation between the learned hyperplanes and the real decision hyperplanes, especially when the size of labeled samples is very small. As a result, pseudo labeling  $\widehat{f}$ is very likely to generate incorrect pseudo labels on unlabeled data points that are close to these biased decision hyperplanes. And our objective now is to optimize the feature representations to reduce the data bias, and finally improve the quality of pseudo labels.

Since we have no labels for $\mathcal{U}$, we cannot directly measure the data bias and thereby reduce it. Yet training bias has some correlations with data bias.
Recall in Section \ref{sec:agent_head}, the task-specific head $h$ is only optimized with clean labeled data, since optimization with incorrect pseudo labels will push the learned hyperplanes in a more biased direction and lead to the training bias. 
Therefore, training bias can be considered as the accumulation of data bias with inappropriate utilization of pseudo labels, which is training algorithm dependent.
And the worst training bias that can be achieved by some self-training methods is a good measure of data bias.
Specifically, the worst training bias corresponds to the worst possible head $h'$ learned by pseudo labeling, such that $h'$ predicts correctly on all the labeled samples $\mathcal{L}$ while making as many mistakes as possible on unlabeled data $\mathcal{U}$,
\begin{equation}
    \label{eq:worst_case}
    h_{\text{worst}}(\psi) = \arg\max_{h'} L_\mathcal{U}(\psi, h', \widehat{f}_{\psi, h}) - L_\mathcal{L}(\psi, h'),
\end{equation}
where the mistakes of $h'$ on unlabeled data are estimated by its discrepancy with the current pseudo labeling function $\widehat{f}$. Equation \ref{eq:worst_case} aims to find the worst-case of task-specific head $h$ that might be learned in the future when trained with pseudo labeling on the current feature generator $\psi$ and the current data sampling.
It is also the \emph{worst hyperplanes}  as shown in Figure \ref{fig:worst_case}(b), which deviates as much as possible from the currently {learned hyperplanes}  while ensuring that all labeled samples are correctly distinguished.
Note that Equation \ref{eq:worst_case} measures the degree of data bias, which depends on the feature representations generated by $\psi$, thus we can adversarially optimize feature generator $\psi$ to indirectly decrease the data bias,
\begin{equation}
    \label{eq:worst_case_objective_psi}
    \min_\psi  L_\mathcal{U}(\psi, h_{\text{worst}}(\psi), \widehat{f}_{\psi, h}) - L_\mathcal{L}(\psi, h_{\text{worst}}(\psi)).
\end{equation}
As shown in Figure \ref{fig:worst_case}(c), Equation \ref{eq:worst_case_objective_psi} encourages the feature of unlabeled samples to be distinguished correctly even by the {worst hyperplanes}, \textit{i.e.}, be generated far away from the {current hyperplanes}, thereby reducing the data bias in feature representations. 

{
\paragraph{Overall loss.}
The final objective of the Debiased Self-Training (DST) approach is to reduce both training bias and data bias.
The overall loss function simultaneously decouples the generation and utilization of pseudo-labels and avoids the worst-case hyperplanes. This is achieved by unifying Equations~\ref{eq:agent}--\ref{eq:worst_case_objective_psi} into a minimax game:
\begin{align}
\label{eq:overall}
\min_{\psi, h, h_{\text{pseudo}}} \max_{h'} L_\mathcal{L}(\psi, h) + L_\mathcal{U}(\psi, h_{\text{pseudo}}, \widehat{f}_{\psi, h}) + \big(    L_\mathcal{U}(\psi, h', \widehat{f}_{\psi, h}) - L_\mathcal{L}(\psi, h')\big).
\end{align}
}

\section{Experiments}
\label{sec:main_experiments}

Following \citep{FixMatch, Dash}, we evaluate Debiased Self-Training (DST) with random initialization on common SSL datasets, including \emph{CIFAR-10} \cite{CIFAR}, \emph{CIFAR-100} \cite{CIFAR}, \emph{SVHN} \cite{SVHN} and \emph{STL-10} \cite{STL10}.
Following \citep{Self-Tuning}, we also 
evaluate DST with both supervised pre-trained models and unsupervised pre-trained models on
 $11$ downstream tasks, including
\textbf{(1)} superordinate-level  object classification: \emph{CIFAR-10} \cite{CIFAR}, \emph{CIFAR-100} \cite{CIFAR}, \emph{Caltech-101} \citep{caltech};
\textbf{(2)} fine-grained object classification: \emph{Food-101} \cite{Food-101}, \emph{CUB-200-2011} \cite{CUB200}, \emph{Stanford Cars} \cite{Stanford-Cars}, \emph{FGVC Aircraft} \cite{Aircraft}, \emph{OxfordIIIT Pets} \cite{Pets}, \emph{Oxford Flowers} \cite{Flowers};
\textbf{(3)} texture classification: \emph{DTD} \cite{DTD};
\textbf{(4)} scene classification: \emph{SUN397} \cite{SUN397}.
The complete training dataset size ranges from $2,040$ to $75, 750$ and the number of classes ranges from $10$ to $397$.
Following \citep{cite:CVPR19DoBetterTransfer}, we report mean accuracy per-class on \emph{Caltech-101}, \emph{FGVC Aircraft}, \emph{OxfordIIIT Pets}, \emph{Oxford Flowers}, and top-1 accuracy for other datasets. 
Following \citep{FixMatch}, we construct a labeled subset with $4$ labels per category
to verify the effectiveness of DST in extremely label-scarce settings.
To make a fair comparison, we keep the labeled subset for each dataset the same throughout our experiments.

For experiments with random initialization, we follow \citep{FixMatch} and adopt Wide ResNet variants \cite{WideResNet}. For experiments with pre-trained models, we adopt ResNet50 \cite{ResNet} with an input size of $224\times224$ and pre-trained on ImageNet \citep{deng_imagenet:_2009}. We adopt MoCo v2 \cite{MoCov2} as unsupervised pre-trained models.
We compare our method with many state-of-the-art SSL methods, including Pseudo Label \citep{pseudo_label}, $\Pi$-Model \cite{Temporal_Ensembling}, Mean Teacher \cite{Mean_Teacher}, {VAT \cite{VAT}, ALI \cite{ALI}, RAT \cite{RAT}}, UDA \cite{UDA}, MixMatch \citep{MixMatch}, ReMixMatch \citep{ReMixMatch}, FixMatch \cite{FixMatch},  Dash \citep{Dash}, Self-Tuning \cite{Self-Tuning}, FlexMatch \citep{FlexMatch} and DebiasMatch \citep{DebiasMatch}.


When training from scratch, we adopt the same hyperparameters as FixMatch \citep{FixMatch}, with learning rate of $0.03$, mini-batch size of $512$. For other experiments, we use SGD with momentum $0.9$ and learning rates in $\{0.001, 0.003, 0.01, 0.03\}$. The mini-batch size is set to $64$ following \citep{Realistic_CVPR}. 
For each image, we first apply random-resize-crop and then  use RandAugment \cite{RandAugment} for strong augmentation $\mathcal{A}$ and  random-horizontal-flip for weak augmentation $\alpha$. More details on hyperparameter selection can be found in Appendix \ref{sec:appendix_hyperparameter}. 
Each experiment is repeated three times
with different random seeds. 
{We have released a benchmark containing both the code for our method and that for all the baselines at \url{https://github.com/thuml/Debiased-Self-Training}.

\begin{figure}[htbp] 
\begin{minipage}[!b]{0.5\textwidth} 
\addtolength{\tabcolsep}{-4pt}
\centering
\scriptsize
\vspace{-10pt}
\tabcaption{
Top-1 accuracy on standard SSL benchmarks
(train from scratch, 4 labels per category).}
\label{tab:train_from_scratch}
\vspace{5pt}
\begin{tabular}{l|cccc|cc}
\toprule
Method             & CIFAR-10 & CIFAR-100 & SVHN & STL-10 & Avg \\ \midrule
Psuedo Label~\cite{pseudo_label}      & 25.4     & 12.6      & 25.3 & 25.3 & 22.2     \\
VAT~\cite{VAT}      &   {25.3} &
  {15.1} &
  {26.1} &
  {25.5} &
  {23.0}      \\
ALI~\cite{ALI}   & {25.9} & {12.4} & {28.5} & {24.1} & {22.7}      \\
RAT~\cite{RAT}   & {33.2} & {20.5} & {52.6} & {30.7} & {34.2}     \\
MixMatch~\cite{MixMatch}          & 52.6     & 32.4      & 57.5 & 45.1 & 46.9     \\
UDA~\cite{UDA}               & 71.0     & 40.7      & 47.4 & 62.6 & 55.4     \\
ReMixMatch~\cite{ReMixMatch}   & 80.9     & 55.7      & 96.6 & 64.0 & 74.3     \\
Dash~\cite{Dash}   & 86.8     & 55.2      & \textbf{97.0} & 64.5 & 75.9     \\ \midrule
FixMatch~\cite{FixMatch}   & 87.2     & 50.6      & 96.5 & 67.1 & 75.4  \\
DST (FixMatch)  & \textbf{89.3}     & \textbf{56.1}      &96.7 & \textbf{71.0} & \textbf{78.3} \\ \midrule
FlexMatch~\cite{FlexMatch}     & 94.7     & 59.5      & 89.6 & 71.3 & 78.8  \\
DST (FlexMatch) & \textbf{95.0}     & \textbf{65.4}      & \textbf{94.2} & \textbf{79.6} & \textbf{83.6}  \\\bottomrule
\end{tabular}
\end{minipage}
\hspace{5pt}
  \begin{minipage}[!b]{0.43\textwidth} 
    \centering
    \vspace{-4pt}
    \caption{Top-1 accuracy on \emph{CIFAR-100} (train from scratch, 4 labels per category). 
    } 
    \vspace{6pt}
        \includegraphics[width=1\textwidth]{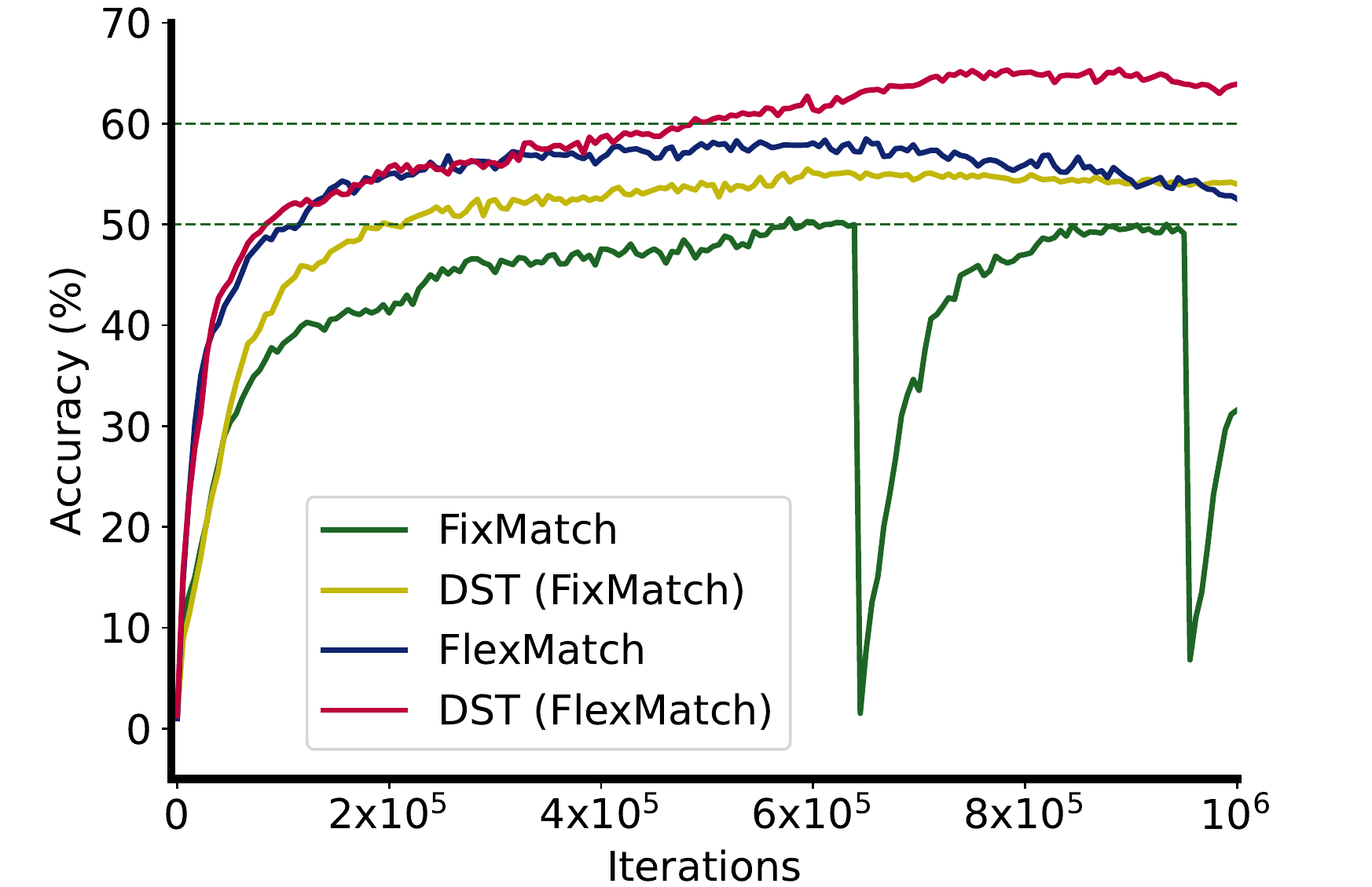}
    \label{fig:scratch_loss_and_acc}
    \vspace{-5pt}
  \end{minipage} 
\end{figure}

\subsection{Main results}
\label{sec:main_result}
Table \ref{tab:train_from_scratch} shows that DST yields consistent improvement on all tasks. On the challenging \emph{CIFAR-100} and \emph{STL-10} tasks, DST boosts the accuracy of FixMatch and FlexMatch by $\textbf{8.3\%}$ and $\textbf{10.7\%}$, respectively.
Figure \ref{fig:scratch_loss_and_acc} depicts the top-1 accuracy during the training procedure on \textit{CIFAR-100}. We observe that
the performance of FixMatch suffers from significant fluctuations during training. In contrast, the accuracy of DST (FixMatch) increases steadily and surpasses the best accuracy of FixMatch by $\textbf{10.9\%}$, relatively. Note that the accuracy of FlexMatch also drops by over $6\%$ in the final stages of training while DST (FlexMatch) suffers from a much smaller drop by reducing erroneous pseudo labels during the self-training process. Besides, DST also improves the performance balance across categories (see Appendix \ref{appdix:performance_balance}).

\subsection{Transfer from a pre-trained model}

\begin{table*}[!bp]
\vspace{-5pt}
\addtolength{\tabcolsep}{-3pt}
\centering
\scriptsize
\caption{Comparison between DST and various baselines (ResNet50, supervised and unsupervised pre-trained, $4$ labels per category). 
\textcolor{red}{$\downarrow$} indicates a performance degradation compared with the baseline. }
\vspace{1pt}
\label{table:main}
\vspace{5pt}
\begin{tabular}{@{}l|l|ccccccccccc|c@{}}
\toprule
& &  \rotatebox{90}{Caltech101} & \rotatebox{90}{CIFAR-10} & \rotatebox{90}{CIFAR-100} & \rotatebox{90}{SUN397} & \rotatebox{90}{DTD} & \rotatebox{90}{Aircraft} & \rotatebox{90}{CUB} & \rotatebox{90}{Flowers} & \rotatebox{90}{Pets} & \rotatebox{90}{Cars} & \rotatebox{90}{Food101} & \rotatebox{90}{Average} \\ \midrule
\multirow{15}{*}{\begin{tabular}[c]{@{}l@{}}\rotatebox{90}{Supervised} \end{tabular}} &~Baseline~ & \textcolor{white}{$\downarrow$}81.4\textcolor{white}{$\downarrow$}   & \textcolor{white}{$\downarrow$}65.2\textcolor{white}{$\downarrow$} & \textcolor{white}{$\downarrow$}48.2\textcolor{white}{$\downarrow$} & \textcolor{white}{$\downarrow$}39.9\textcolor{white}{$\downarrow$} & \textcolor{white}{$\downarrow$}47.7\textcolor{white}{$\downarrow$} & \textcolor{white}{$\downarrow$}25.4\textcolor{white}{$\downarrow$} & \textcolor{white}{$\downarrow$}46.5\textcolor{white}{$\downarrow$} & \textcolor{white}{$\downarrow$}85.2\textcolor{white}{$\downarrow$} & \textcolor{white}{$\downarrow$}78.1\textcolor{white}{$\downarrow$} & \textcolor{white}{$\downarrow$}33.3\textcolor{white}{$\downarrow$} & \textcolor{white}{$\downarrow$}33.8\textcolor{white}{$\downarrow$} &~53.2 \\
 &~Pseudo Label~\cite{pseudo_label}~ & 86.3    & 83.3    & 54.7   & 41.0   & 50.2   & 27.2   & 54.3   & 92.3   & 87.8   & 41.4   & 38.0   &~59.7 \\
 &~$\Pi$-Model~\cite{Temporal_Ensembling}~ & 83.5   & 73.1   & 49.2   & \textcolor{white}{$\downarrow$}39.7\textcolor{red}{$\downarrow$}   & 50.3   & \textcolor{white}{$\downarrow$}24.3\textcolor{red}{$\downarrow$}   & 47.1   & 90.7   & 82.2   & \textcolor{white}{$\downarrow$}30.9\textcolor{red}{$\downarrow$}   & 33.9   &~55.0 \\
 &~Mean Teacher~\cite{Mean_Teacher}~ & 83.7   & 82.1   & 56.0   & \textcolor{white}{$\downarrow$}37.9\textcolor{red}{$\downarrow$}   & 51.6   & 30.7   & 49.6   & 91.0   & 82.8   & 39.1   & 40.3   &~58.6 \\
 & {~VAT~\cite{VAT}~} & {84.1} &
  {72.2} &
  {48.8} &
  \textcolor{white}{$\downarrow$}{39.5}\textcolor{red}{$\downarrow$} &
  {50.6} &
  {25.9} &
  {48.1} &
  {89.4} &
  {81.8} &
  \textcolor{white}{$\downarrow$}{32.4}\textcolor{red}{$\downarrow$} &
  {36.7} &
  {~55.4}  \\
  & {~ALI~\cite{ALI}~} & {82.2} &
  {69.5} &
  \textcolor{white}{$\downarrow$}{46.3}\textcolor{red}{$\downarrow$} &
  \textcolor{white}{$\downarrow$}{36.4}\textcolor{red}{$\downarrow$} &
  {50.5} &
  \textcolor{white}{$\downarrow$}{21.3}\textcolor{red}{$\downarrow$} &
  \textcolor{white}{$\downarrow$}{42.5}\textcolor{red}{$\downarrow$} &
  \textcolor{white}{$\downarrow$}{82.9}\textcolor{red}{$\downarrow$} &
  \textcolor{white}{$\downarrow$}{77.4}\textcolor{red}{$\downarrow$} &
  \textcolor{white}{$\downarrow$}{29.8}\textcolor{red}{$\downarrow$} &
  \textcolor{white}{$\downarrow$}{31.7}\textcolor{red}{$\downarrow$} &
  {~51.9}
\\
& {~RAT~\cite{RAT}~} & {84.0} &
  {81.8} &
  {55.4} &
  \textcolor{white}{$\downarrow$}{39.0}\textcolor{red}{$\downarrow$} &
  {49.1} &
  {31.6} &
  {50.0} &
  {89.9} &
  {84.1} &
  {37.9} &
  {38.4} &
  {~58.3}
\\
 & {~MixMatch~\cite{MixMatch}~} & {85.4} &
  {82.8} &
  {53.5} &
  {41.8} &
  {50.1} &
  \textcolor{white}{$\downarrow$}{24.7}\textcolor{red}{$\downarrow$} &
  {51.7} &
  {91.5} &
  {83.3} &
  {42.5} &
  {38.2} &
  {~58.7}  \\
 &~UDA~\cite{UDA}~ & 85.8   & 83.6   & 54.7   & 41.3   & 49.0   & 27.1   & 52.1   & 92.0   & 83.1   & 45.6   & 41.7   &~59.6 \\
 &~FixMatch~\cite{FixMatch}~ & 86.3   & 84.6   & 53.1   & 41.3   & 48.6   & \textcolor{white}{$\downarrow$}25.2\textcolor{red}{$\downarrow$}   & 52.3   & 93.2   & 83.7   & 46.4   & 37.1   &~59.3 \\
 &~Self-Tuning~\cite{Self-Tuning}~ & 87.2   & 76.0   & 57.1   & 41.8   & 50.7   & 35.2   & 58.9   & 92.6   & 86.6   & 58.3   & 41.9   &~62.4 \\
 &~FlexMatch~\cite{FlexMatch}~ & 87.1 & 89.0 & 63.4 & 48.3 & 52.5 & 34.0 & 54.9 & 94.5 & 88.3 & 57.5 & 49.5 &~65.4 \\
 &~DebiasMatch~\cite{DebiasMatch}~ & 88.6 & 91.0 & 65.7 & 46.6 & 52.4 & 37.5 & 58.6 & 95.6 & 86.4 & 60.5 & 53.5 &~66.9 \\  \cmidrule(l){2-14} 
 &~DST (FixMatch)~ & 89.6  & 94.9   & 70.4  & 48.1  & 53.5  & 43.2  & 68.7  & 94.8  & 89.8  & 71.0  & \textbf{58.5}  &~71.1 \\
 & {~DST (FlexMatch)}~ & {\textbf{90.6}} &
  {\textbf{95.9}} &
  {\textbf{71.2}} &
  {\textbf{49.8}} &
  {\textbf{56.2}} &
  {\textbf{44.5}} &
  {\textbf{70.5}} &
  {\textbf{95.8}} &
  {\textbf{90.4}} &
  {\textbf{72.7}} &
  {57.1} &
  {~\textbf{72.2}} \\
\midrule
\multirow{15}{*}{\begin{tabular}[c]{@{}l@{}}\rotatebox{90}{Unsupervised}\end{tabular}} &~Baseline~ & 79.5   & 66.6   & 46.5  & 38.1  & 47.9  & 28.7  & 37.5  & 87.7  & 60.0  & 38.1  & 32.9  &~51.2 \\
 &~Pseudo Label~\cite{pseudo_label}~ & 86.2  & 70.8 & 49.8 & 38.6 & 50.0 & \textcolor{white}{$\downarrow$}26.6\textcolor{red}{$\downarrow$}  & 41.8 & 93.0 & 68.4 &
 \textcolor{white}{$\downarrow$}37.3\textcolor{red}{$\downarrow$} & \textcolor{white}{$\downarrow$}32.8\textcolor{red}{$\downarrow$} &~54.1 \\
 &~$\Pi$-Model~\cite{Temporal_Ensembling}~ & 80.1 & 76.2 & \textcolor{white}{$\downarrow$}44.8\textcolor{red}{$\downarrow$} & \textcolor{white}{$\downarrow$}37.8\textcolor{red}{$\downarrow$} & 50.0 & \textcolor{white}{$\downarrow$}23.5\textcolor{red}{$\downarrow$} & \textcolor{white}{$\downarrow$}31.6\textcolor{red}{$\downarrow$} & 93.1 & 62.8 & \textcolor{white}{$\downarrow$}25.6\textcolor{red}{$\downarrow$} & \textcolor{white}{$\downarrow$}30.4\textcolor{red}{$\downarrow$} &~50.5 \\
 &~Mean Teacher~\cite{Mean_Teacher}~ & 80.4 & 80.8 & 51.3 & \textcolor{white}{$\downarrow$}34.2\textcolor{red}{$\downarrow$} & 48.8 & 33.8 & 41.6 & 92.9 & 67.0 & 50.5 & 39.1 &~56.4 \\
 & {~VAT~\cite{VAT}~} & {79.9} &
  {73.8} &
  \textcolor{white}{$\downarrow$}{45.1}\textcolor{red}{$\downarrow$} &
  {38.3} &
  {49.2} &
  \textcolor{white}{$\downarrow$}{24.2}\textcolor{red}{$\downarrow$} &
  \textcolor{white}{$\downarrow$}{36.4}\textcolor{red}{$\downarrow$} &
  {92.4} &
  {61.7} &
  \textcolor{white}{$\downarrow$}{29.9\textcolor{red}{$\downarrow$}} &
  {33.1} &
  {~51.3}  \\
 & {~ALI~\cite{ALI}~} & \textcolor{white}{$\downarrow$}{76.4}\textcolor{red}{$\downarrow$} &
  {69.2} &
  \textcolor{white}{$\downarrow$}{44.4}\textcolor{red}{$\downarrow$} &
  \textcolor{white}{$\downarrow$}{34.9}\textcolor{red}{$\downarrow$} &
  {50.1} &
  \textcolor{white}{$\downarrow$}{22.2}\textcolor{red}{$\downarrow$} &
  \textcolor{white}{$\downarrow$}{33.8}\textcolor{red}{$\downarrow$} &
  \textcolor{white}{$\downarrow$}{84.9}\textcolor{red}{$\downarrow$} &
  \textcolor{white}{$\downarrow$}{59.6}\textcolor{red}{$\downarrow$} &
  \textcolor{white}{$\downarrow$}{33.1}\textcolor{red}{$\downarrow$} &
  \textcolor{white}{$\downarrow$}{31.0}\textcolor{red}{$\downarrow$} &
  {~49.1}
\\
& {~RAT~\cite{RAT}~} & {80.9} &
  {79.5} &
  {52.4} &
  \textcolor{white}{$\downarrow$}{37.0}\textcolor{red}{$\downarrow$} &
  {50.4} &
  {30.1} &
  {40.7} &
  {91.8} &
  {70.5} &
  {47.9} &
  {35.6} &
  {~56.1}
\\
 & {~MixMatch~\cite{MixMatch}~} & {84.1} &
  {81.5} &
  {51.7} &
  {38.4} &
  \textcolor{white}{$\downarrow$}{47.0}\textcolor{red}{$\downarrow$} &
  {31.7} &
  {39.8} &
  {93.5} &
  {66.4} &
  {47.1} &
  {34.6} &
  {~56.0}  \\
 &~UDA~\cite{UDA}~ & 85.0 & 87.4 & 53.6 & 42.3 & \textcolor{white}{$\downarrow$}46.2\textcolor{red}{$\downarrow$} & 35.7 & 41.4 & 94.1 & 69.3 & 51.5 & 39.3 &~58.7 \\
 &~FixMatch~\cite{FixMatch} & 83.1 & 82.2 & 51.4 & 39.2 & \textcolor{white}{$\downarrow$}43.9\textcolor{red}{$\downarrow$} & 30.1 & \textcolor{white}{$\downarrow$}36.8\textcolor{red}{$\downarrow$} & 94.3 & 65.7 & 48.6 & 36.8 &~55.6 \\
 &~Self-Tuning~\cite{Self-Tuning}~ & 81.6 & \textcolor{white}{$\downarrow$}63.6\textcolor{red}{$\downarrow$} & 47.8 & 38.8 & \textcolor{white}{$\downarrow$}45.5\textcolor{red}{$\downarrow$} & 31.4 & 41.6 & 91.0 & 66.9 & 52.0 & 34.0 &~54.0 \\
 &~FlexMatch~\cite{FlexMatch}~ & 86.4 & 96.7 & 60.2 & 45.3 & 53.9 & 42.0 & 49.2 & 95.8 & 72.9 & 69.0 & 37.5 &~64.4 \\
  &~DebiasMatch~\cite{DebiasMatch}~ & 86.4 & 96.3 & 66.3 & 44.5 & 53.9 & 44.8 & 51.2 & 95.4 & 70.9 & 72.5 & 53.6 &~66.9 \\ \cmidrule(l){2-14} 
 &~DST (FixMatch)~ & 90.1 & 95.0 & 68.2 & 46.8 & 54.2 & \textbf{47.7} & 53.6 & 95.6 & \textbf{75.4} & 72.0 & \textbf{57.1} &~68.7 \\
 & {~DST (FlexMatch)}~ & {\textbf{90.4}} &
  {\textbf{96.9}} &
  {\textbf{68.9}} &
  {\textbf{48.8}} &
  {\textbf{55.9}} &
  {47.3} &
  {\textbf{55.2}} &
  {\textbf{96.4}} &
  {75.1} &
  {\textbf{74.6}} &
  {56.9} &
  {~\textbf{69.7}} \\
 \bottomrule
\end{tabular}
\end{table*}

\textbf{{Supervisied pre-training.}} Table \ref{table:main} reveals that typical self-training methods, e.g. FixMatch, lead to relatively mild improvements with supervised pre-trained models, which is consistent with previous findings \cite{Realistic_CVPR, Self-Tuning}.
In contrast, incorporating DST into FixMatch significantly boosts the performance and surpasses FixMatch by $\textbf{19.9\%}$ on all datasets.
With a pre-trained model, self-training has better training stability.
Yet once the performance degradation occurs, the process is also irreversible (Appendix \ref{appdix:training_stablity}), partly due to the catastrophic forgetting of pre-trained representation. Also, self-training suffers from a more severe performance imbalance across classes (Appendix \ref{appdix:performance_balance}). DST effectively tackles these issues, indicating the importance of reducing bias.

\textbf{{Unsupervised pre-training.}} Table \ref{table:main} shows that with unsupervised pre-trained models, more methods suffer from performance degradation after self-training on the unlabeled data. 
The difficulty comes from that the unsupervised pre-training task has a larger task discrepancy with the downstream classification tasks than the supervised pre-training task. Thus, the representations learned by unsupervised pre-trained models usually exhibit stronger data bias, and inappropriate usage of pseudo labels will lead to rapid accumulation errors and increase the training bias. By eliminating training bias and reducing data bias, DST brings improvement on all datasets and relatively outperforms FixMatch by $\textbf{23.5\%}$ on average, superior to FlexMatch and DebiasMatch in $9$ and $10$ tasks, respectively.


\subsection{Ablation studies}
We examine the design of our method on \emph{CIFAR-100} in Table \ref{table:ablation} and have the following findings.
\textbf{(1)} Compared with \textit{Mutual Learning} \cite{mutual, MMT}, where two heads provide pseudo labels to each other, 
the independent mechanism in our method where one head is only responsible for generating pseudo labels and the other head only uses them for self-training can better reduce the training bias.
\textbf{(2)} 
A nonlinear pseudo head is always better than a linear pseudo head. We conjecture that nonlinear projection can reduce the degeneration of representation with biased pseudo labels.
\textbf{(3)} The worst-case estimation of pseudo labeling improves the performance by large margins.

\begin{table*}[htbp]
\vspace{-5pt}
\addtolength{\tabcolsep}{0pt}
\centering
\scriptsize
\caption{Ablation study on \textit{CIFAR-100} with different pre-trained models (4 labels per category). 
}
\label{table:ablation}
\vspace{5pt}
\begin{tabular}{@{}l|cccc|cc@{}}
\toprule
\multirow{2}{*}{Method}                                                                  & \multirow{2}{*}{\begin{tabular}[c]{@{}c@{}}Multiple \\ Heads\end{tabular}} & \multirow{2}{*}{\begin{tabular}[c]{@{}c@{}}Linear \\ Pseudo Head\end{tabular}}  & \multirow{2}{*}{\begin{tabular}[c]{@{}c@{}}Nonlinear \\ Pseudo Head\end{tabular}} & \multirow{2}{*}{\begin{tabular}[c]{@{}c@{}}Worst Case\\ Estimation\end{tabular}} & Supervised & Unsupervised \\
& & & & & Pre-training & Pre-training \\ \midrule
FixMatch & & & & & 53.1 & 51.4 \\
Mutual Learning & \checkmark & & & & 53.4 & 52.5 \\
DST w/o worst & \checkmark & \checkmark & & & 58.2 & 59.0  \\
DST w/o worst & \checkmark &  & \checkmark & & 60.6 & 60.9  \\
DST & \checkmark &  & \checkmark & \checkmark & \textbf{70.4} & \textbf{68.2} \\ \bottomrule
\end{tabular}
\end{table*}

\subsection{Analysis}
\label{sec:experiments_analysis}
To further investigate how DST improves pseudo labeling and self-training performance, we conduct some analysis on \emph{CIFAR-100}. For simplicity, we only give the results with supervised pre-trained models. More comparisons can be found in Appendix \ref{appendix:analysis_pseudo}.

\textbf{{DST improves both the quantity and quality of pseudo labels.}} As shown in Figures \ref{fig:raito} and \ref{fig:accuracy}, FixMatch exploits unlabeled data \emph{aggressively}, on average producing more than $70\%$ pseudo labels during training. But the cost is that the accuracy of pseudo labels continues to drop, eventually falling below $60\%$, which is consistent with our motivation in Section \ref{sec:analysis} that inappropriate utilization of pseudo labels will in turn enlarges the training bias.
On the contrary, the accuracy of pseudo labels in DST suffers from a smaller drop. Rather, it keeps rising afterward and exceeds $70\%$ throughout the training. Besides, DST generates more pseudo labels in the later stages of training.


\begin{figure}[htbp]
    \centering
    \subfigure[Quantity]{
        \includegraphics[width=0.22\textwidth]{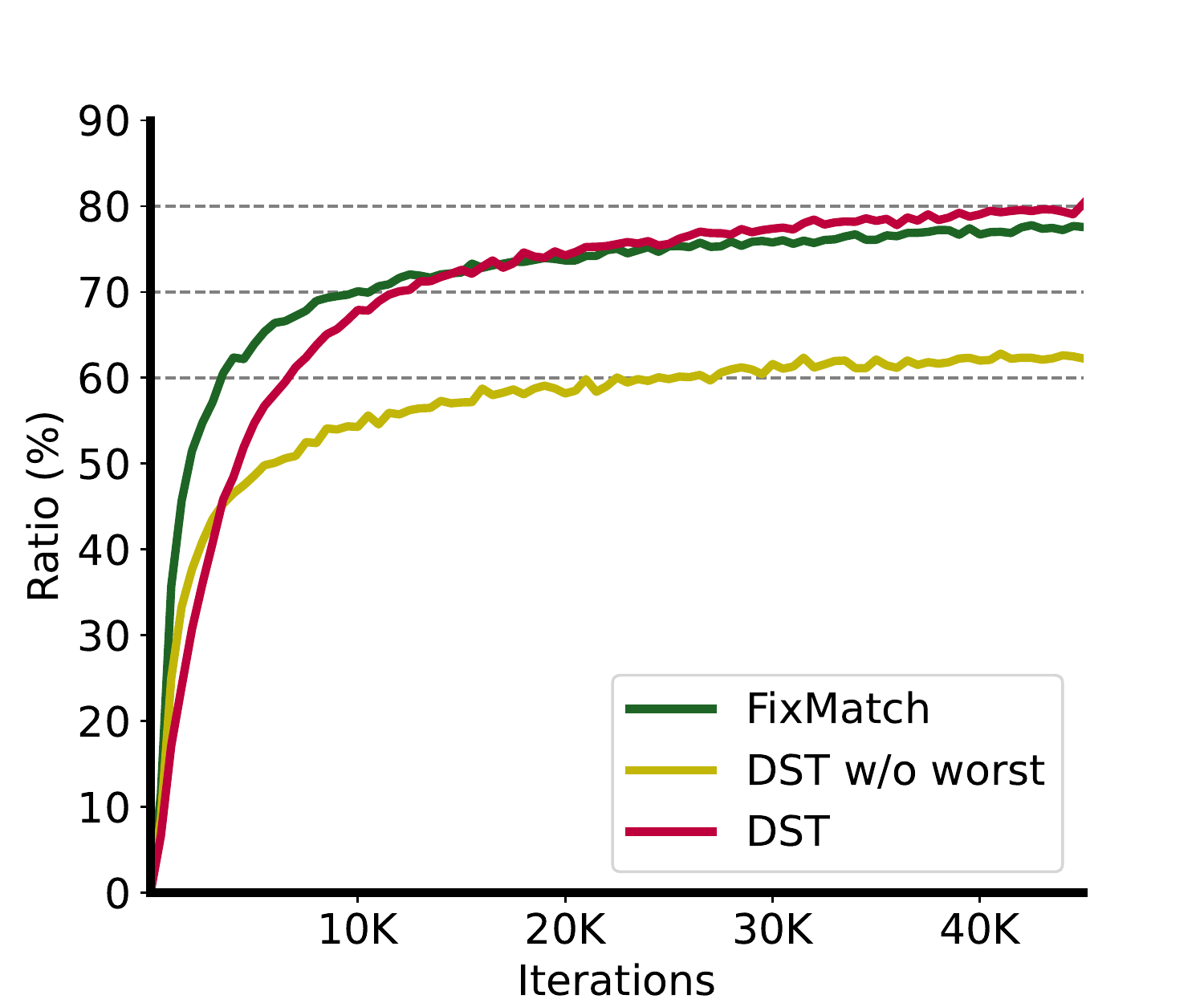}
        \label{fig:raito}
    }
    \subfigure[Quality]{
        \includegraphics[width=0.22\textwidth]{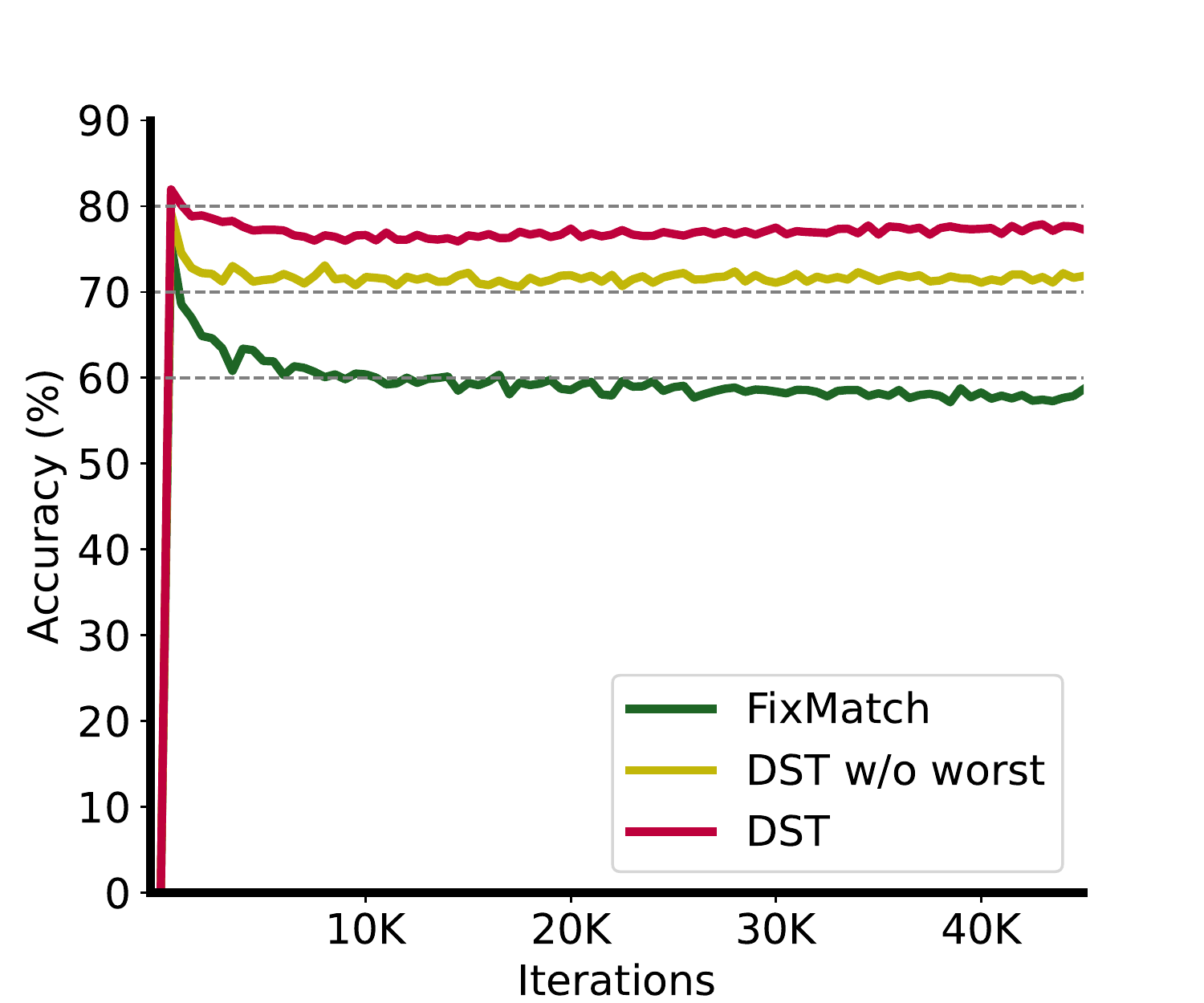}
        \label{fig:accuracy}
    }
    \subfigure[Quantity of bad classes]{
        \includegraphics[width=0.22\textwidth]{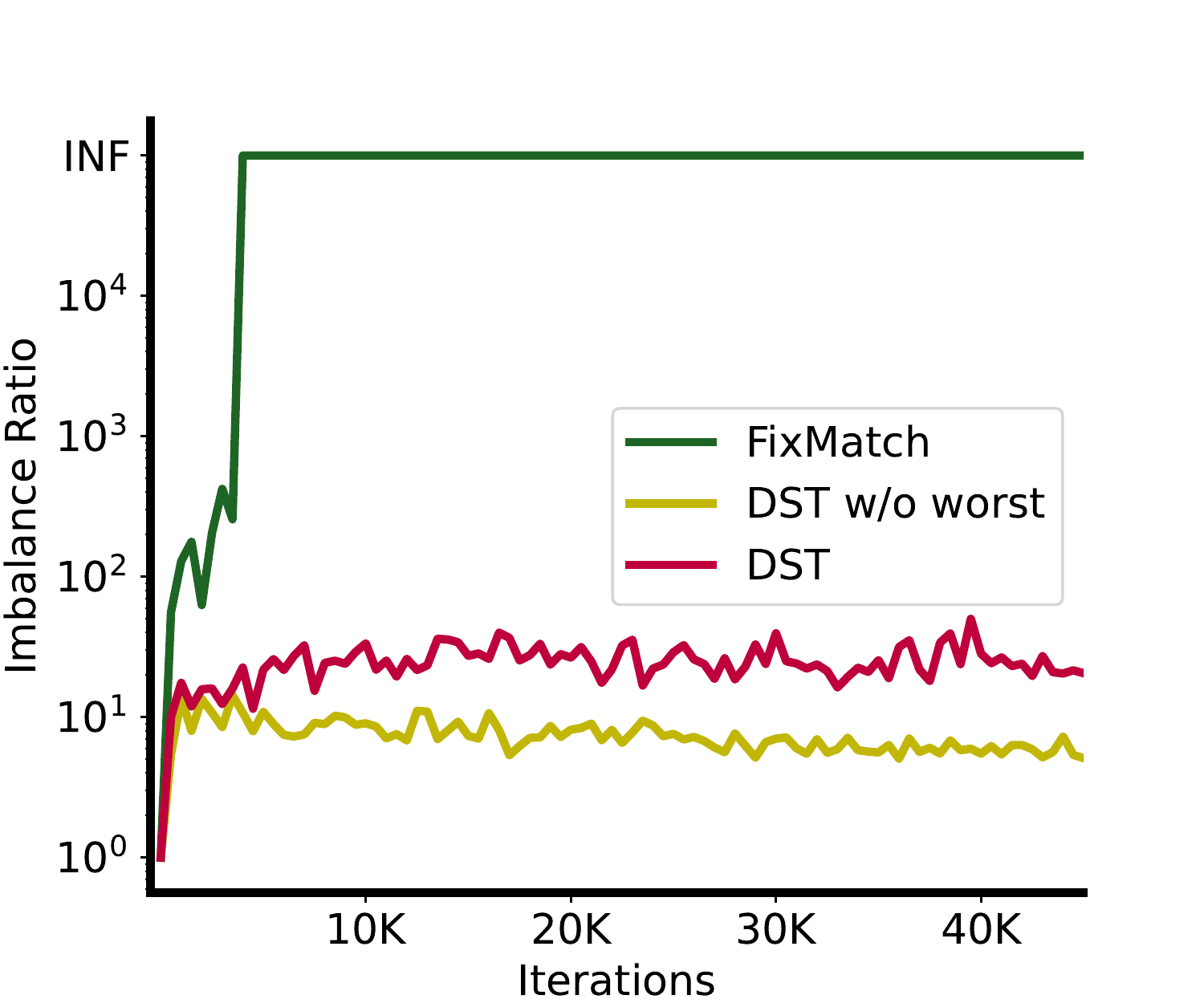}
        \label{fig:imbalance}
    }
    \subfigure[Quality of bad classes]{
        \includegraphics[width=0.22\textwidth]{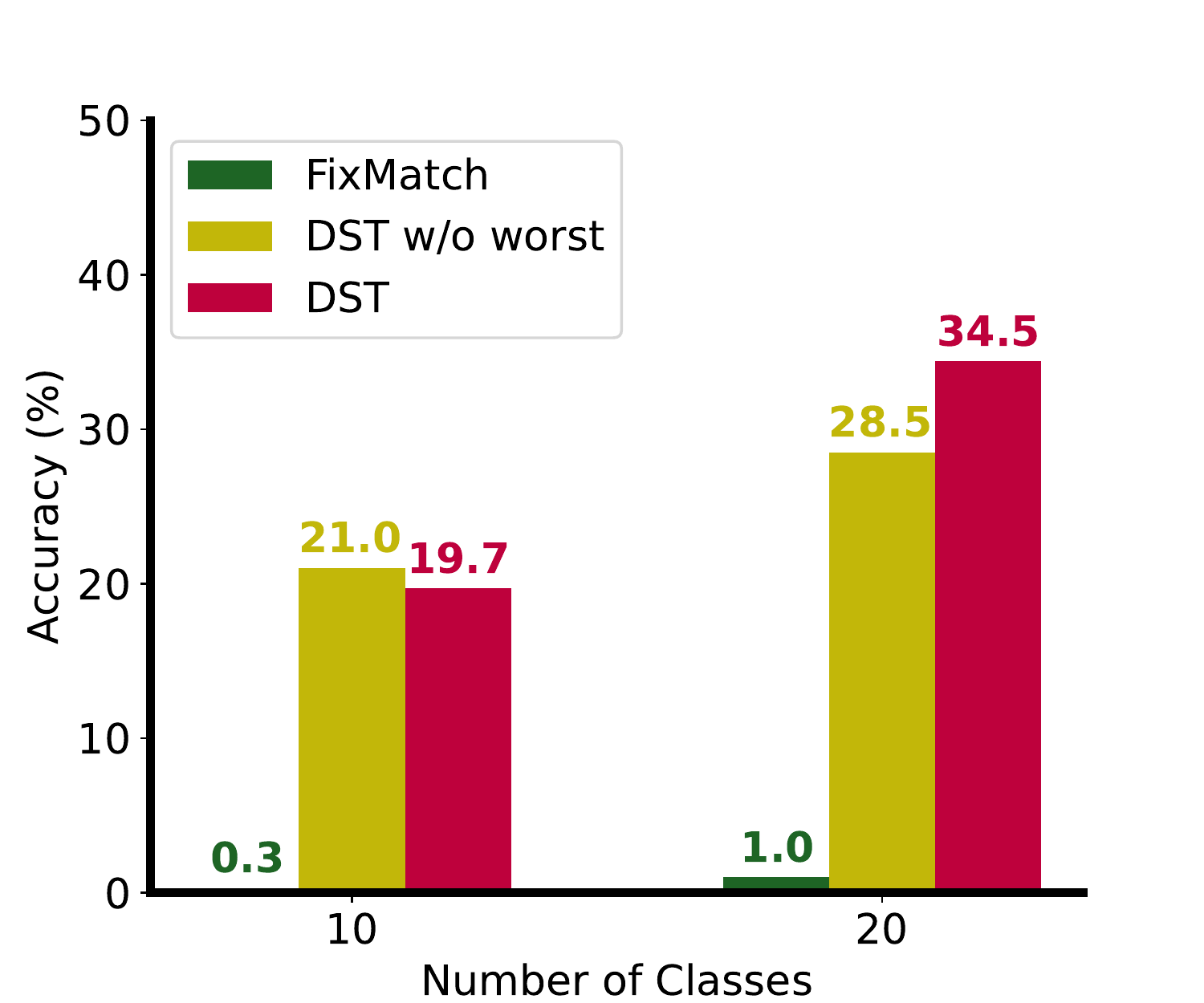}
        \label{fig:worst}
    }
    \caption{The quantity and quality of pseudo labels  on \emph{CIFAR-100} (ResNet50, supervised pre-trained).}
\end{figure}

\textbf{{DST generates better pseudo labels for poorly-behaved classes.}}
To measure the quantity of pseudo labels on poorly-behaved classes, we calculate the class imbalance ratio $I$ on a class-balanced validation set,
$
    I={{\max}_cN(c)}/{{\min}_{c'} N(c')}
$,
where $N(c)$ denotes the number of predictions that fall into category $c$.
As shown in Figure \ref{fig:imbalance}, the class imbalance ratio of FixMatch rises rapidly and reaches infinity after $5000$ iterations, indicating that the model completely ignores those poorly-learned classes.
To measure the quality of pseudo labels on poorly-behaved classes,
we calculate the average accuracy of $10$ or $20$ worst-behaved classes in 
Figure \ref{fig:worst}. The average accuracy on the worst $20$ classes of FixMatch is only \textbf{1.0\%}. By reducing training bias with the pseudo head and data bias with the worst-case estimation, the average accuracy balloons to \textbf{28.5\%} and \textbf{34.5\%}, respectively. 

\subsection{Convergence and computation cost of the min-max optimization}
We optimize $\psi$ and $h'$ with stochastic gradient descent alternatively. The optimization can be viewed as an alternative form of GAN \cite{GAN}. Figure~\ref{fig:convergence} shows that the \textcolor[rgb]{0, 0.7, 0}{worst-case error rate} of $h'$ and \textcolor{red}{worst-case loss} in Equation \ref{eq:worst_case_objective_psi} first increase ($h'$ dominates), and then gradually decrease and converge ($\psi$ dominates). When training $1000k$ iterations on \emph{CIFAR-100} using $4$ 2080 Ti GPUs, FixMatch takes $104$ hours while DST takes $111$ hours, only a $7\%$ increase in time. Note that DST introduces no additional computation cost during inference.

\begin{figure}[htbp] 
\begin{minipage}[!b]{0.43\textwidth} 
    \centering
\caption{Empirical error rate and loss (\textit{CIFAR-100}).}
\vspace{0pt}
\includegraphics[width=1\textwidth]{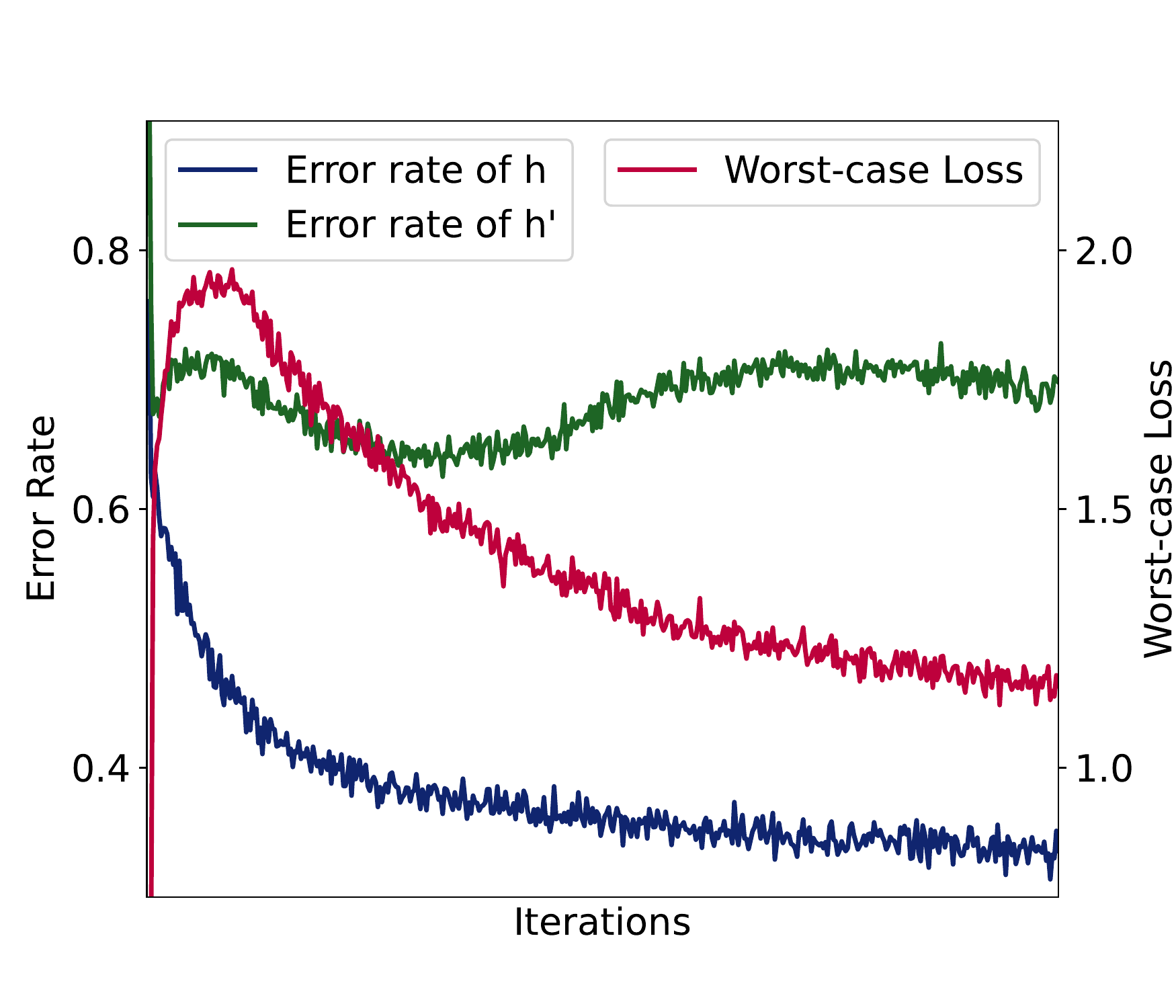}
\label{fig:convergence}
\end{minipage} 
\hspace{5pt}
\begin{minipage}[!b]{0.5\textwidth} 
\vspace{-20pt}
\addtolength{\tabcolsep}{-1pt}
\caption{DST as a general add-on  on \textit{CIFAR-100}. }
\vspace{5pt}
\label{table:framework}
\centering
\scriptsize
\begin{tabular}{@{}ll|cc|cc@{}}
\toprule
\multicolumn{2}{l|}{Pre-training} & \multicolumn{2}{c|}{Supervised} & \multicolumn{2}{c}{Unsupervised} \\ \midrule
\multicolumn{2}{l|}{Label Amount} & 400 & 1000 & 400 & 1000 \\ \midrule
\multicolumn{1}{l|}{\multirow{2}{*}{\begin{tabular}[c]{@{}l@{}}Mean \\ Teacher \end{tabular}}} & Base & 56.0 & 67.0 & 51.3 & 63.5 \\
\multicolumn{1}{l|}{} & DST & \textbf{62.7} & \textbf{70.7} & \textbf{60.7} & \textbf{69.3} \\ \midrule
\multicolumn{1}{l|}{\multirow{2}{*}{\begin{tabular}[c]{@{}l@{}}Noisy \\ Student\end{tabular}}} & Base & 52.8 & 64.3 & 55.6 & 65.8 \\
\multicolumn{1}{l|}{} & DST & \textbf{68.9} & \textbf{74.8} & \textbf{66.6} & \textbf{75.2} \\ \midrule
\multicolumn{1}{l|}{\multirow{2}{*}{{DivideMix}}} & {Base} & {55.8} & {67.5} & {53.6} & {64.9} \\
\multicolumn{1}{l|}{} & {DST} & {\textbf{69.1}} & {\textbf{75.1}} & {\textbf{65.0}} & {\textbf{74.2}} \\ \midrule
\multicolumn{1}{l|}{\multirow{2}{*}{FixMatch}} & Base & 53.1 & 67.8 & 51.4 & 64.2 \\
\multicolumn{1}{l|}{} & DST & \textbf{70.4}  & \textbf{75.6} & \textbf{68.2}  & \textbf{76.8} \\ \midrule
\multicolumn{1}{l|}{\multirow{2}{*}{FlexMatch}} & Base & 63.4 & 71.2 & 60.2 & 71.1 \\
\multicolumn{1}{l|}{} & DST & \textbf{71.2}  & \textbf{77.3} & \textbf{68.9}  & \textbf{77.5} \\ \bottomrule
\end{tabular}
\end{minipage}

\vspace{-10pt}
\end{figure}

\subsection{DST as a general add-on}
\label{sec:framework}
We incorporate DST into several representative self-training methods, including FixMatch \cite{FixMatch}, Mean Teacher \cite{Mean_Teacher}, Noisy Student \cite{Noisy_Student}, {DivideMix \cite{DivideMix}} and FlexMatch \cite{FlexMatch}.
Implementation details of DST versions of these methods can be found in Appendix \ref{appendix:framework}. 
Table \ref{table:framework} compares the original  and DST versions of these methods on \emph{CIFAR-100} with both supervised pre-trained and unsupervised pre-trained models. 
Results show that the proposed DST yields large improvement on all these self-training methods, indicating that self-training bias widely exists in existing vanilla or sophisticated self-training methods and DST can serve as a universal add-on to reduce the bias.

\section{Conclusion}
To mitigate the requirement for labeled data, pseudo labels are widely used on the unlabeled data, yet they suffer from severe confirmation bias. 
In this paper, we systematically delved into the bias issues 
and present Debiased Self-Training (DST), a novel approach to decrease bias in self-training. Experimentally, {DST} achieves state-of-the-art performance on $13$ semi-supervised learning tasks and can serve as a universal and beneficial add-on for existing self-training methods.

\section*{Acknowledgements}
This work was supported by the National Key Research and Development Plan (2021YFB1715200), National Natural Science Foundation of China (62022050 and 62021002), Beijing Nova Program (Z201100006820041), BNRist Innovation Fund (BNR2021RC01002), and Kuaishou Research Fund.

\bibliographystyle{plain}
\bibliography{main}

\begin{thebibliography}{10}

\bibitem{pseudoLabel2019}
Eric Arazo, Diego Ortego, Paul Albert, Noel~E O'Connor, and Kevin McGuinness.
\newblock Pseudo-labeling and confirmation bias in deep semi-supervised
  learning.
\newblock In {\em IJCNN}, 2020.

\bibitem{PAWS}
Mahmoud Assran, Mathilde Caron, Ishan Misra, Piotr Bojanowski, Armand Joulin,
  Nicolas Ballas, and Michael Rabbat.
\newblock Semi-supervised learning of visual features by non-parametrically
  predicting view assignments with support samples.
\newblock In {\em ICCV}, 2021.

\bibitem{ReMixMatch}
David Berthelot, Nicholas Carlini, Ekin~D Cubuk, Alex Kurakin, Kihyuk Sohn, Han
  Zhang, and Colin Raffel.
\newblock Remixmatch: Semi-supervised learning with distribution alignment and
  augmentation anchoring.
\newblock In {\em ICLR}, 2020.

\bibitem{MixMatch}
David Berthelot, Nicholas Carlini, Ian Goodfellow, Nicolas Papernot, Avital
  Oliver, and Colin Raffel.
\newblock Mixmatch: A holistic approach to semi-supervised learning.
\newblock In {\em NeurIPS}, 2019.

\bibitem{CoTraining}
Avrim Blum and Tom Mitchell.
\newblock Combining labeled and unlabeled data with co-training.
\newblock In {\em Proceedings of the eleventh annual conference on
  Computational learning theory}, 1998.

\bibitem{Food-101}
Lukas Bossard, Matthieu Guillaumin, and Luc Van~Gool.
\newblock Food-101--mining discriminative components with random forests.
\newblock In {\em ECCV}, 2014.

\bibitem{Simclrv2}
Ting Chen, Simon Kornblith, Kevin Swersky, Mohammad Norouzi, and Geoffrey
  Hinton.
\newblock Big self-supervised models are strong semi-supervised learners.
\newblock In {\em NeurIPS}, 2020.

\bibitem{MoCov2}
Xinlei Chen, Haoqi Fan, Ross Girshick, and Kaiming He.
\newblock Improved baselines with momentum contrastive learning.
\newblock {\em arXiv preprint arXiv:2003.04297}, 2020.

\bibitem{DTD}
Mircea Cimpoi, Subhransu Maji, Iasonas Kokkinos, Sammy Mohamed, and Andrea
  Vedaldi.
\newblock Describing textures in the wild.
\newblock In {\em CVPR}, 2014.

\bibitem{STL10}
Adam Coates, Andrew Ng, and Honglak Lee.
\newblock An analysis of single-layer networks in unsupervised feature
  learning.
\newblock In {\em AISTATS}, 2011.

\bibitem{RandAugment}
Ekin~D Cubuk, Barret Zoph, Jonathon Shlens, and Quoc~V Le.
\newblock Randaugment: Practical automated data augmentation with a reduced
  search space.
\newblock In {\em CVPR}, 2020.

\bibitem{GoodSSLBadGAN}
Zihang Dai, Zhilin Yang, Fan Yang, William~W Cohen, and Russ~R Salakhutdinov.
\newblock Good semi-supervised learning that requires a bad gan.
\newblock In {\em NeurIPS}, 2017.

\bibitem{deng_imagenet:_2009}
Jia Deng, Wei Dong, Richard Socher, Li-Jia Li, Kai Li, and Li~Fei-Fei.
\newblock Imagenet: {A} large-scale hierarchical image database.
\newblock In {\em CVPR}, 2009.

\bibitem{cite:NAACL19BERT}
Jacob Devlin, Ming-Wei Chang, Kenton Lee, and Kristina Toutanova.
\newblock Bert: Pre-training of deep bidirectional transformers for language
  understanding.
\newblock In {\em NAACL}, 2019.

\bibitem{ALI}
Vincent Dumoulin, Ishmael Belghazi, Ben Poole, Olivier Mastropietro, Alex Lamb,
  Martin Arjovsky, and Aaron Courville.
\newblock Adversarially learned inference.
\newblock In {\em ICLR}, 2017.

\bibitem{caltech}
Li~Fei-Fei, R.~Fergus, and P.~Perona.
\newblock Learning generative visual models from few training examples: An
  incremental bayesian approach tested on 101 object categories.
\newblock In {\em CVPR}, 2004.

\bibitem{MMT}
Yixiao Ge, Dapeng Chen, and Hongsheng Li.
\newblock Mutual mean-teaching: Pseudo label refinery for unsupervised domain
  adaptation on person re-identification.
\newblock In {\em ICLR}, 2020.

\bibitem{GAN}
Ian Goodfellow, Jean Pouget-Abadie, Mehdi Mirza, Bing Xu, David Warde-Farley,
  Sherjil Ozair, Aaron Courville, and Yoshua Bengio.
\newblock Generative adversarial nets.
\newblock In {\em NeurIPS}, 2014.

\bibitem{adversarial_samples}
Ian~J Goodfellow, Jonathon Shlens, and Christian Szegedy.
\newblock Explaining and harnessing adversarial examples.
\newblock In {\em ICLR}, 2015.

\bibitem{entropy_minimization}
Yves Grandvalet and Yoshua Bengio.
\newblock Semi-supervised learning by entropy minimization.
\newblock In {\em NeurIPS}, 2005.

\bibitem{cite:CVPR20MoCo}
Kaiming He, Haoqi Fan, Yuxin Wu, Saining Xie, and Ross Girshick.
\newblock Momentum contrast for unsupervised visual representation learning.
\newblock In {\em CVPR}, 2020.

\bibitem{ResNet}
Kaiming He, Xiangyu Zhang, Shaoqing Ren, and Jian Sun.
\newblock Deep residual learning for image recognition.
\newblock In {\em CVPR}, 2016.

\bibitem{label_propagation}
Ahmet Iscen, Giorgos Tolias, Yannis Avrithis, and Ondrej Chum.
\newblock Label propagation for deep semi-supervised learning.
\newblock In {\em CVPR}, 2019.

\bibitem{jiang2022transferability}
Junguang Jiang, Yang Shu, Jianmin Wang, and Mingsheng Long.
\newblock Transferability in deep learning: A survey, 2022.

\bibitem{catastrophic_forgetting}
James Kirkpatrick, Razvan Pascanu, Neil Rabinowitz, Joel Veness, Guillaume
  Desjardins, Andrei~A. Rusu, Kieran Milan, John Quan, Tiago Ramalho, Agnieszka
  Grabska-Barwinska, Demis Hassabis, Claudia Clopath, Dharshan Kumaran, and
  Raia Hadsell.
\newblock Overcoming catastrophic forgetting in neural networks.
\newblock {\em Proceedings of the National Academy of Sciences}, 2017.

\bibitem{cite:CVPR19DoBetterTransfer}
Simon Kornblith, Jonathon Shlens, and Quoc~V Le.
\newblock Do better imagenet models transfer better?
\newblock In {\em CVPR}, 2019.

\bibitem{Stanford-Cars}
Jonathan Krause, Jia Deng, Michael Stark, and Li~Fei-Fei.
\newblock Collecting a large-scale dataset of fine-grained cars.
\newblock In {\em FGVC}, 2013.

\bibitem{CIFAR}
Alex Krizhevsky, Geoffrey Hinton, et~al.
\newblock Learning multiple layers of features from tiny images.
\newblock {\em Technical report, University of Toronto}, 2009.

\bibitem{Temporal_Ensembling}
Samuli Laine and Timo Aila.
\newblock Temporal ensembling for semi-supervised learning.
\newblock In {\em ICLR}, 2017.

\bibitem{pseudo_label}
Dong-Hyun Lee.
\newblock Pseudo-label: The simple and efficient semi-supervised learning
  method for deep neural networks.
\newblock In {\em ICML}, 2013.

\bibitem{DivideMix}
Junnan Li, Richard Socher, and Steven~CH Hoi.
\newblock Dividemix: Learning with noisy labels as semi-supervised learning.
\newblock In {\em ICLR}, 2020.

\bibitem{CoMatch}
Junnan Li, Caiming Xiong, and Steven~CH Hoi.
\newblock Comatch: Semi-supervised learning with contrastive graph
  regularization.
\newblock In {\em ICCV}, 2021.

\bibitem{Aircraft}
Subhransu Maji, Esa Rahtu, Juho Kannala, Matthew Blaschko, and Andrea Vedaldi.
\newblock Fine-grained visual classification of aircraft.
\newblock {\em arXiv preprint arXiv:1306.5151}, 2013.

\bibitem{VAT}
Takeru Miyato, Shin-ichi Maeda, Masanori Koyama, and Shin Ishii.
\newblock Virtual adversarial training: a regularization method for supervised
  and semi-supervised learning.
\newblock In {\em TPAMI}, 2018.

\bibitem{SVHN}
Yuval Netzer, Tao Wang, Adam Coates, Alessandro Bissacco, Bo~Wu, and Andrew~Y
  Ng.
\newblock Reading digits in natural images with unsupervised feature learning.
\newblock In {\em NeurIPS}, 2011.

\bibitem{Flowers}
Maria-Elena Nilsback and Andrew Zisserman.
\newblock Automated flower classification over a large number of classes.
\newblock In {\em ICVGIP}, 2008.

\bibitem{GANSSL}
Augustus Odena.
\newblock Semi-supervised learning with generative adversarial networks.
\newblock {\em arXiv preprint arXiv:1606.01583}, 2016.

\bibitem{DASO}
Youngtaek Oh, Dong-Jin Kim, and In~So Kweon.
\newblock Daso: Distribution-aware semantics-oriented pseudo-label for
  imbalanced semi-supervised learning.
\newblock In {\em CVPR}, 2022.

\bibitem{VAdD}
Sungrae Park, JunKeon Park, Su-Jin Shin, and Il-Chul Moon.
\newblock Adversarial dropout for supervised and semi-supervised learning.
\newblock In {\em AAAI}, 2018.

\bibitem{Pets}
Omkar~M Parkhi, Andrea Vedaldi, Andrew Zisserman, and CV~Jawahar.
\newblock Cats and dogs.
\newblock In {\em CVPR}, 2012.

\bibitem{Pytorch}
Adam Paszke, Sam Gross, Francisco Massa, Adam Lerer, James Bradbury, Gregory
  Chanan, Trevor Killeen, Zeming Lin, Natalia Gimelshein, Luca Antiga, et~al.
\newblock Pytorch: An imperative style, high-performance deep learning library.
\newblock In {\em NeurIPS}, 2019.

\bibitem{meta_pseudo_labels}
Hieu Pham, Zihang Dai, Qizhe Xie, and Quoc~V Le.
\newblock Meta pseudo labels.
\newblock In {\em CVPR}, 2021.

\bibitem{CLIP}
Alec Radford, Jong~Wook Kim, Chris Hallacy, Aditya Ramesh, Gabriel Goh,
  Sandhini Agarwal, Girish Sastry, Amanda Askell, Pamela Mishkin, Jack Clark,
  et~al.
\newblock Learning transferable visual models from natural language
  supervision.
\newblock In {\em ICML}, 2021.

\bibitem{Defense}
Mamshad~Nayeem Rizve, Kevin Duarte, Yogesh~S Rawat, and Mubarak Shah.
\newblock In defense of pseudo-labeling: An uncertainty-aware pseudo-label
  selection framework for semi-supervised learning.
\newblock In {\em ICLR}, 2021.

\bibitem{self_training_wacv}
Chuck Rosenberg, Martial Hebert, and Henry Schneiderman.
\newblock Semi-supervised self-training of object detection models.
\newblock In {\em WACV}, 2005.

\bibitem{MTTriTraining}
Sebastian Ruder and Barbara Plank.
\newblock Strong baselines for neural semi-supervised learning under domain
  shift.
\newblock In {\em ACL}, 2018.

\bibitem{ImproveGANs}
Tim Salimans, Ian Goodfellow, Wojciech Zaremba, Vicki Cheung, Alec Radford, and
  Xi~Chen.
\newblock Improved techniques for training gans.
\newblock In {\em NeurIPS}, 2016.

\bibitem{transductive}
Weiwei Shi, Yihong Gong, Chris Ding, Zhiheng~MaXiaoyu Tao, and Nanning Zheng.
\newblock Transductive semi-supervised deep learning using min-max features.
\newblock In {\em ECCV}, 2018.

\bibitem{FixMatch}
Kihyuk Sohn, David Berthelot, Chun-Liang Li, Zizhao Zhang, Nicholas Carlini,
  Ekin~D Cubuk, Alex Kurakin, Han Zhang, and Colin Raffel.
\newblock Fixmatch: Simplifying semi-supervised learning with consistency and
  confidence.
\newblock In {\em NeurIPS}, 2020.

\bibitem{Dropout}
Nitish Srivastava, Geoffrey Hinton, Alex Krizhevsky, Ilya Sutskever, and Ruslan
  Salakhutdinov.
\newblock Dropout: a simple way to prevent neural networks from overfitting.
\newblock In {\em ICML}, 2014.

\bibitem{Realistic_CVPR}
Jong-Chyi Su, Zezhou Cheng, and Subhransu Maji.
\newblock A realistic evaluation of semi-supervised learning for fine-grained
  classification.
\newblock In {\em CVPR}, 2021.

\bibitem{RAT}
Teppei Suzuki and Ikuro Sato.
\newblock Adversarial transformations for semi-supervised learning.
\newblock In {\em AAAI}, 2020.

\bibitem{Mean_Teacher}
Antti Tarvainen and Harri Valpola.
\newblock Mean teachers are better role models: Weight-averaged consistency
  targets improve semi-supervised deep learning results.
\newblock In {\em NeurIPS}, 2017.

\bibitem{CUB200}
Catherine Wah, Steve Branson, Peter Welinder, Pietro Perona, and Serge
  Belongie.
\newblock The caltech-ucsd birds-200-2011 dataset.
\newblock {\em Technical Report CNS-TR-2011-001, California Institute of
  Technology}, 2011.

\bibitem{Self-Tuning}
Ximei Wang, Jinghan Gao, Mingsheng Long, and Jianmin Wang.
\newblock Self-tuning for data-efficient deep learning.
\newblock In {\em ICML}, 2021.

\bibitem{DebiasMatch}
Xudong Wang, Zhirong Wu, Long Lian, and Stella~X Yu.
\newblock Debiased learning from naturally imbalanced pseudo-labels for
  zero-shot and semi-supervised learning.
\newblock In {\em CVPR}, 2022.

\bibitem{self_training_theory}
Colin Wei, Kendrick Shen, Yining Chen, and Tengyu Ma.
\newblock Theoretical analysis of self-training with deep networks on unlabeled
  data.
\newblock In {\em ICLR}, 2021.

\bibitem{SUN397}
Jianxiong Xiao, James Hays, Krista~A Ehinger, Aude Oliva, and Antonio Torralba.
\newblock Sun database: Large-scale scene recognition from abbey to zoo.
\newblock In {\em CVPR}, 2010.

\bibitem{UDA}
Qizhe Xie, Zihang Dai, Eduard Hovy, Minh-Thang Luong, and Quoc~V Le.
\newblock Unsupervised data augmentation for consistency training.
\newblock In {\em NeurIPS}, 2020.

\bibitem{Noisy_Student}
Qizhe Xie, Minh-Thang Luong, Eduard Hovy, and Quoc~V Le.
\newblock Self-training with noisy student improves imagenet classification.
\newblock In {\em CVPR}, 2020.

\bibitem{Dash}
Yi~Xu, Lei Shang, Jinxing Ye, Qi~Qian, Yu-Feng Li, Baigui Sun, Hao Li, and Rong
  Jin.
\newblock Dash: Semi-supervised learning with dynamic thresholding.
\newblock In {\em ICML}, 2021.

\bibitem{self_training_acl}
David Yarowsky.
\newblock Unsupervised word sense disambiguation rivaling supervised methods.
\newblock In {\em ACL}, 1995.

\bibitem{WideResNet}
Sergey Zagoruyko and Nikos Komodakis.
\newblock Wide residual networks.
\newblock In {\em BMVC}, 2016.

\bibitem{FlexMatch}
Bowen Zhang, Yidong Wang, Wenxin Hou, Hao Wu, Jindong Wang, Manabu Okumura, and
  Takahiro Shinozaki.
\newblock Flexmatch: Boosting semi-supervised learning with curriculum pseudo
  labeling.
\newblock In {\em NeurIPS}, 2021.

\bibitem{mutual}
Ying Zhang, Tao Xiang, Timothy~M Hospedales, and Huchuan Lu.
\newblock Deep mutual learning.
\newblock In {\em CVPR}, 2018.

\bibitem{cite:ICLR21InstanceTransfer}
Nanxuan Zhao, Zhirong Wu, Rynson W.~H. Lau, and Stephen Lin.
\newblock What makes instance discrimination good for transfer learning?
\newblock In {\em ICLR}, 2021.

\bibitem{place}
Bolei Zhou, Agata Lapedriza, Aditya Khosla, Aude Oliva, and Antonio Torralba.
\newblock Places: A 10 million image database for scene recognition.
\newblock In {\em TPAMI}, 2018.

\end{thebibliography}

\newpage
\appendix

\section{Implementation Details}
\label{appendix:implementation_details}

Our code is based on PyTorch \cite{Pytorch}. The following are the implementation details of our experiments. {We have released the code for our method and that for all the baselines at \url{https://github.com/thuml/Debiased-Self-Training}}. 

\subsection{Architecture}
The architectures of different classifier heads are as follows.
For nonlinear heads, we adopt Dropout \cite{Dropout} to alleviate over-fitting.
\begin{itemize}
    \item Linear main head: \texttt{Linear-Softmax};
    \item Nonlinear pseudo head: \texttt{Linear-ReLU-Dropout-Linear-Softmax};
    \item Worst-case head: \texttt{Linear-ReLU-Dropout-Linear-Softmax}.
\end{itemize}

\subsection{Hyperparameters}
\label{sec:appendix_hyperparameter}
For experiments \textit{without} pre-trained models, we use Wide ResNet-28-2 \cite{WideResNet} for \emph{CIFAR-10} and \emph{SVHN}, WRN-28-8 for \emph{CIFAR-100}, WRN-37-2 for \emph{STL-10} and adopt the same hyperparameters as FixMatch \cite{FixMatch}. Specifically, we use learning rate of $0.03$, mini-batch size of $512$ ($64$ for labeled data, $448$ for unlabeled data), weight decay in $\{0.0005, 0.001\}$, unlabeled loss weight $\lambda=1.0$ and train for $1000k$ iterations.
For our method, we set the projection dimension of the pseudo head and worst-case head to $2\times n_{\text{embedding}}$, where $n_{\text{embedding}}$ denotes the dimension of backbone network output.

For experiments \textit{with} pre-trained models, we use SGD with momentum of $0.9$. We choose weight decay in $\{0.0005, 0.001\}$, learning rates in $\{0.001, 0.003, 0.01, 0.03\}$. We train for $40k$ iterations and use the cosine learning rate schedule. The mini-batch size is set to $64$. Besides, we tune the following algorithm-specific hyperparameters.

\textbf{$\Pi$-Model}. We search unlabeled loss weight $\lambda$ in $\{0.1, 0.3, 1.0, 3.0\}$, warm-up iterations of unlabeled loss in $\{5\times10^3, 10^4\}$. 

\textbf{Mean Teacher}.  We fix the exponential moving average hyperparameter $\alpha$ to $0.999$. We search unlabeled loss weight $\lambda$ in $\{0.1, 0.3, 1.0, 3.0\}$, warm-up iterations of unlabeled loss in $\{5\times10^3, 10^4\}$.

\textbf{Pseudo Label}.
We search confidence threshold $\tau$ in $\{0.7, 0.8, 0.9, 0.95\}$, unlabeled loss weight $\lambda$ in $\{0.1, 0.3, 1.0, 3.0\}$. 

\textbf{FixMatch}. The same as Pseudo Label.

\textbf{UDA}.
We search temperature in $\{0.1, 0.5, 1.0, 2.0\}$, unlabeled loss weight $\lambda$ in $\{0.1, 0.3, 1.0, 3.0\}$. 

\textbf{Noisy Student}.
We search temperature in $\{0.1, 0.5, 1.0, 2.0\}$, unlabeled loss weight $\lambda$ in $\{0.1, 0.3, 1.0, 3.0\}$. We iterate $3$ rounds, excluding the round that trains with only labeled data. The final performance is reported.

\textbf{Self-Tuning}.
We try queue size in $\{24, 32\}$ and use temperature $0.07$, projection dimension $1024$, same as the original paper \cite{Self-Tuning}.

\textbf{FlexMatch.} 
The same as Pseudo Label. Besides, we find it better to turn off threshold warm-up when using pre-trained models.

\textbf{DebiasMatch.} We search confidence threshold $\tau$ in $\{0.7, 0.8, 0.9, 0.95\}$, unlabeled loss weight $\lambda$ in $\{0.1, 0.3, 1.0, 3.0\}$, debias factor in $\{0.1, 0.3, 1.0, 3.0\}$ and fix momentum to $0.999$. For a fair comparison, we do not exploit additional supervision from pre-trained CLIP models \cite{CLIP}.

\textbf{DST}.
We set the confidence threshold to $0.7$ by default. We fix the projection dimension of the pseudo head and the worst-case head to $2048$. The trade-off hyperparameter $\lambda$ is set to $1$.

\subsection{DST as a general add-on to previous self-training methods}
\begin{figure*}[htbp]
    \centering
    \includegraphics[width=1\textwidth]{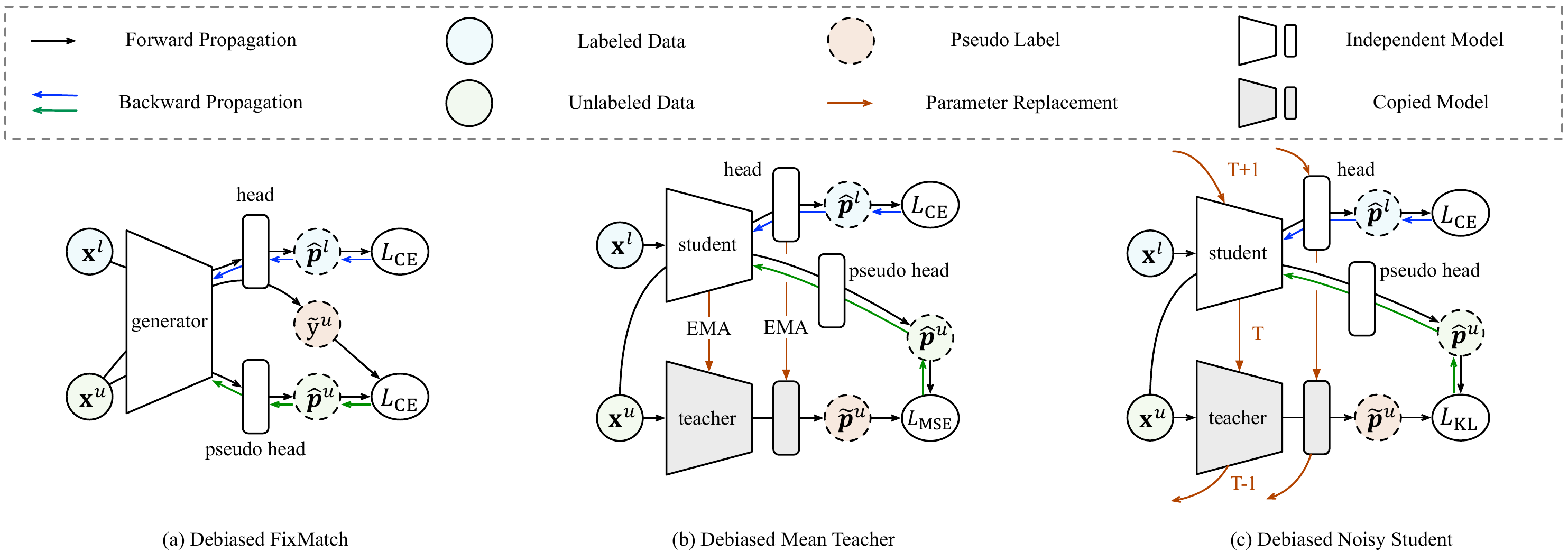}
    \vspace{-10pt}
    \caption{Illustrations on how different Debiased self-training methods generate and utilize pseudo labels.}
    \label{fig:debiased_family}
\end{figure*}

\label{appendix:framework}
In this section, we will illustrate how to incorporate DST into $4$ typical self-training methods, including FixMatch, FlexMatch, Mean Teacher, and Noisy Student. 
We will mainly focus on the modification to the generation and utilization of pseudo labels, and omit the introduction of the worst-case heads since they are the same across different self-training methods.

\textbf{Debiased FixMatch. } 
As shown in Figure \ref{fig:debiased_family}(a), the pseudo labels on the unlabeled data are generated by the main head $h$ and utilized by the pseudo head $h_{\text{pseudo}}$. The main head $h$ is only trained on the clean labeled data.

\textbf{Debiased FlexMatch. } 
The same as Debiased FixMatch. The learning status of each category is estimated by the main head $h$.

\textbf{Debiased Mean Teacher.}
As shown in Figure \ref{fig:debiased_family}(b),  the pseudo labels on the unlabeled data are generated by the exponential moving average of the main head $h$ and utilized by the pseudo head $h_{\text{pseudo}}$. The main head $h$ is only trained on the clean labeled data.

\textbf{Debiased Noisy Student.}
As shown in Figure \ref{fig:debiased_family}(c), 
the pseudo labels  on the unlabeled data are generated by  the head $h$ of previous round $T-1$ and utilized by the pseudo head $h_{\text{pseudo}}$. The main head $h$ is only trained on the clean labeled data.

\section{More Experimental Results}
\label{appendix:more_results}

\subsection{Experiments on training stability}
\label{appdix:training_stablity}
We further explore the training stability of FixMatch when using pre-trained models on various tasks. Figure \ref{fig:failure_case} illustrates several failure cases of FixMatch.

\begin{figure}[h]
    \centering
    \subfigure[\emph{Aircraft} (supervised pre-trained)]{
        \includegraphics[width=0.3\textwidth]{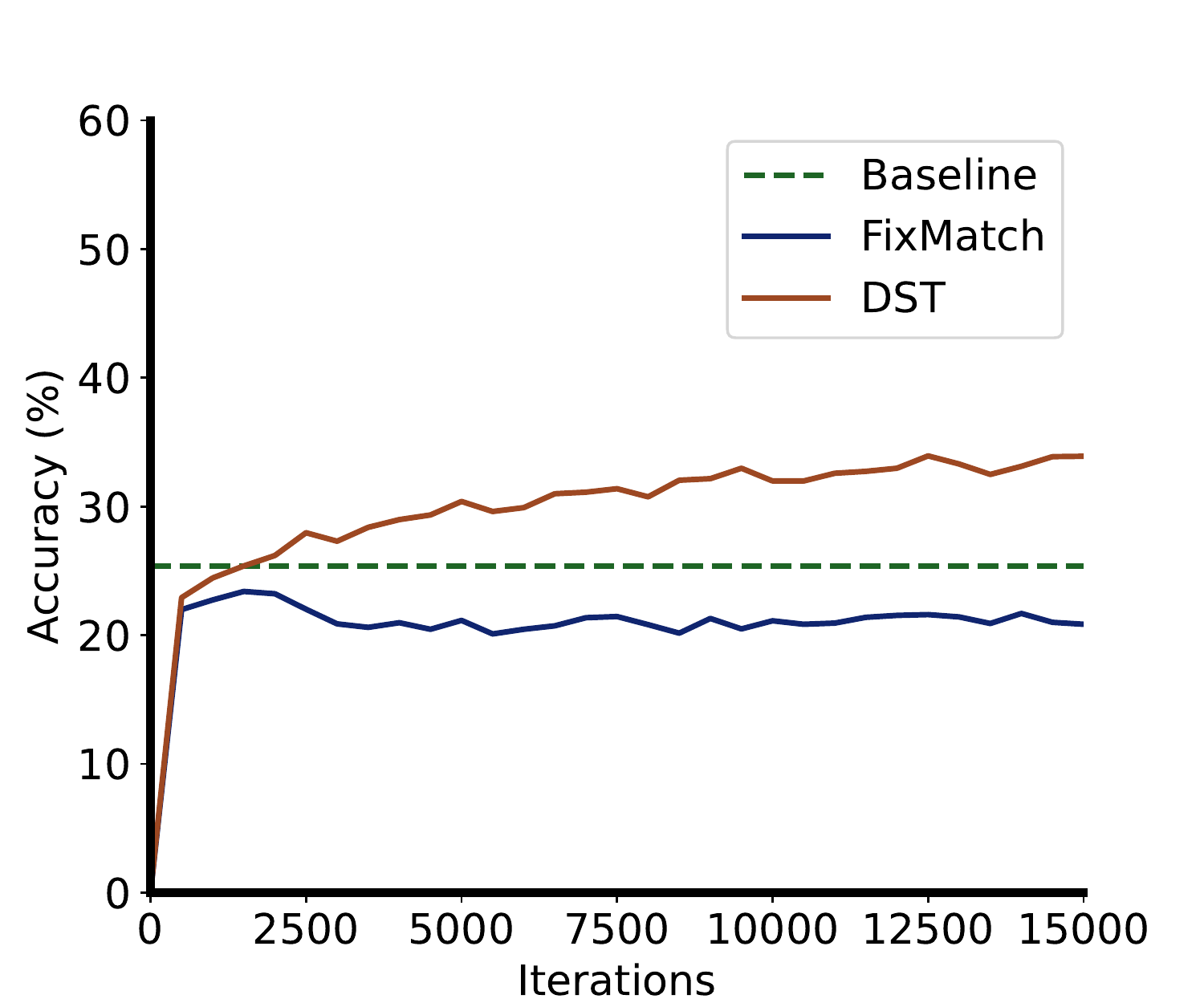}
        \label{appendix:aircraft}
    }
    \subfigure[\emph{CUB} (unsupervised pre-trained)]{
        \includegraphics[width=0.3\textwidth]{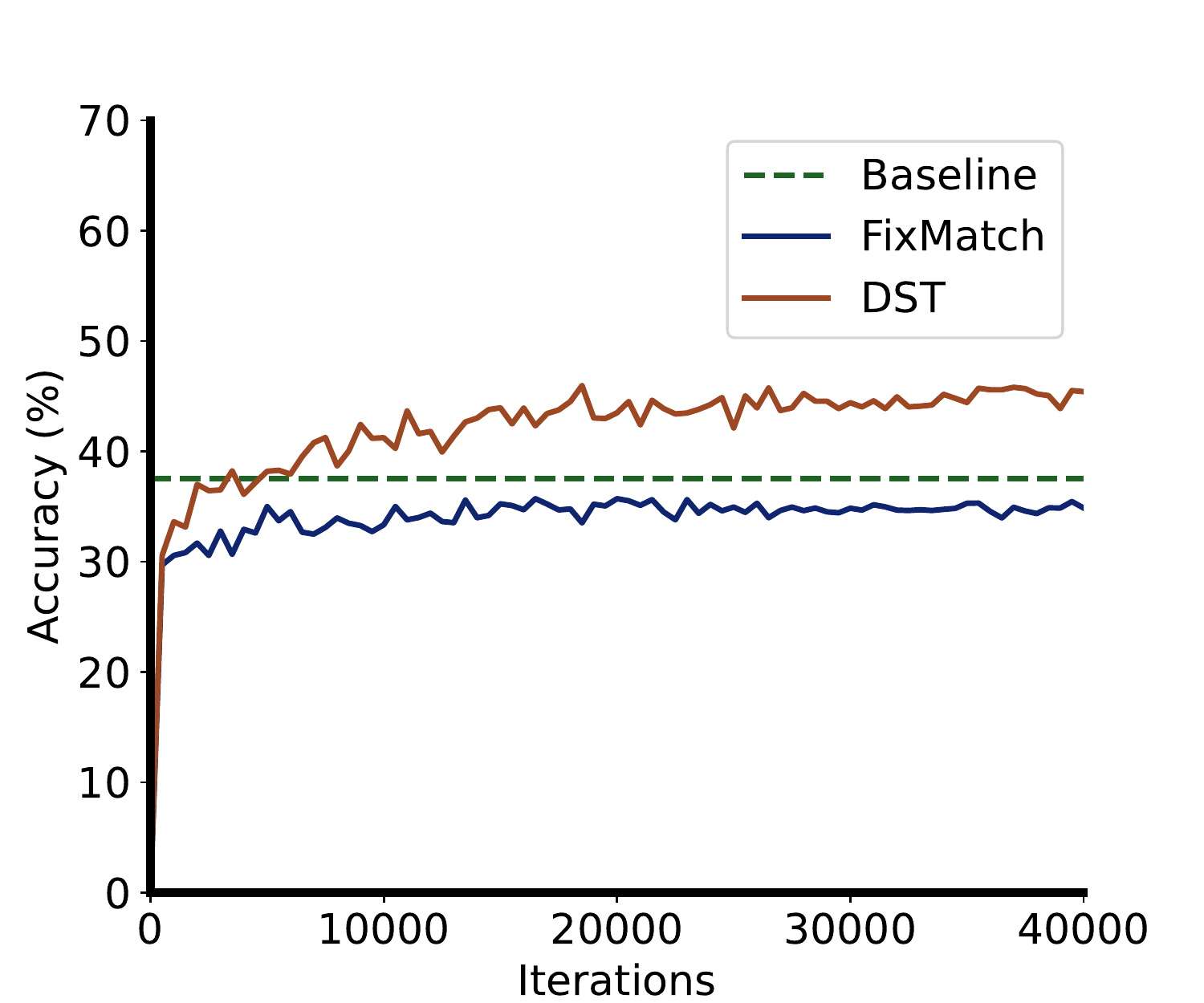}
        \label{appendix:cub}
    }
    \subfigure[\emph{Food-101} (unsupervised pre-trained)]{
        \includegraphics[width=0.3\textwidth]{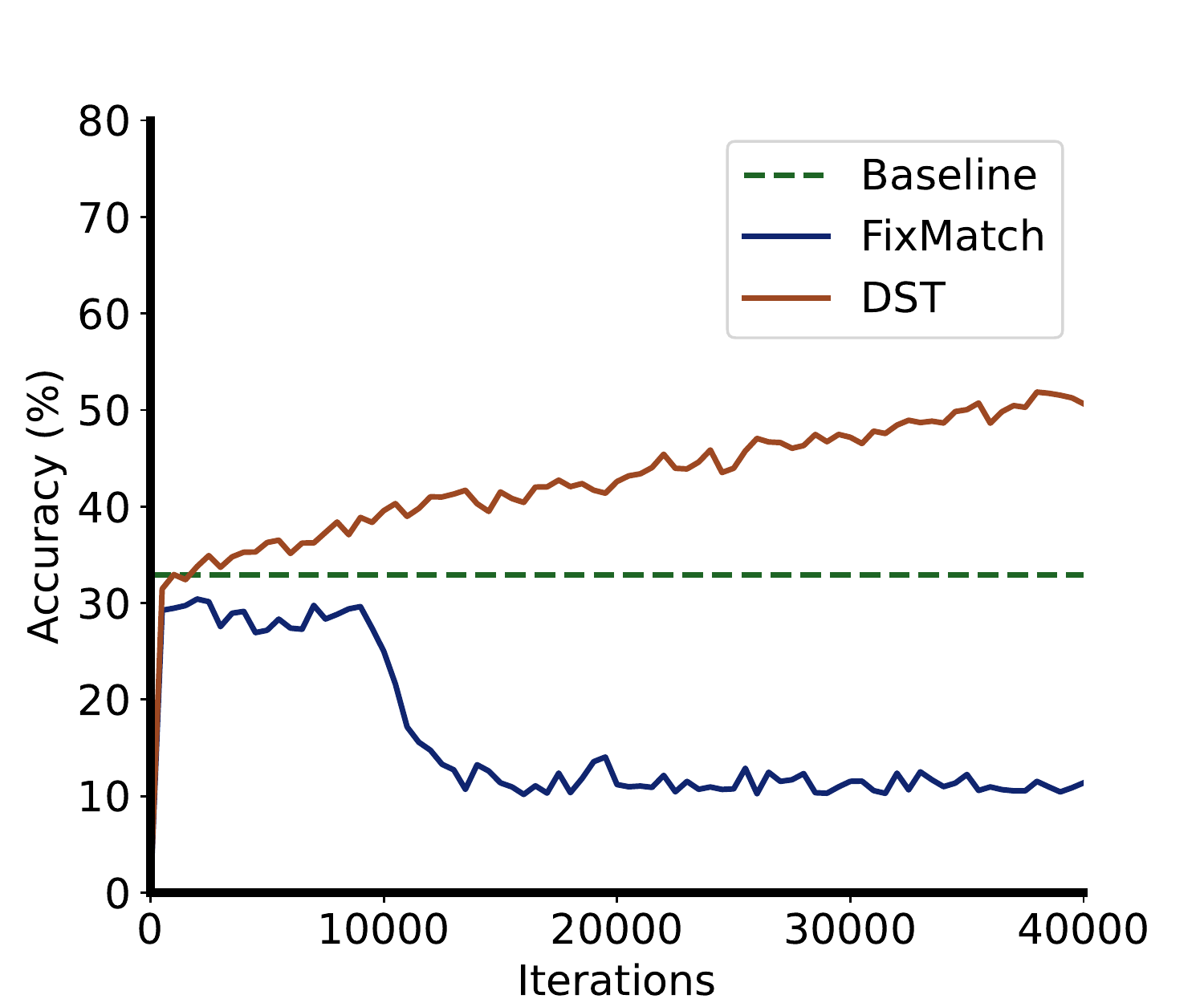}
        \label{appendix:food}
    }
    \caption{Failure cases of FixMatch with confidence threshold $0.7$ (ResNet50). }
    \label{fig:failure_case}
\end{figure}

Figures \ref{appendix:aircraft} and \ref{appendix:cub} show \{when the performance of the pre-trained models declines, it cannot be recovered later}. Note that we try confidence threshold in $\{0.7, 0.8, 0.9, 0.95\}$ and get similar results. 

Figure \ref{appendix:food} demonstrates a complete failure case of FixMatch. With unsupervised pre-trained models and a confidence threshold of $0.7$, there can be a lot of noise in pseudo labels and thus the performance of FixMatch declines severely. Note this result is not the entry we report in Table \ref{table:main} (threshold is set to $0.9$ for this dataset). Instead, we aim to show that {DST improves the training stability when there is much noise}.

\subsection{Experiments on performance balance between categories}
\label{appdix:performance_balance}
\begin{figure}[!ht]
    \centering
    \subfigure[FixMatch]{
        \includegraphics[width=0.3\textwidth]{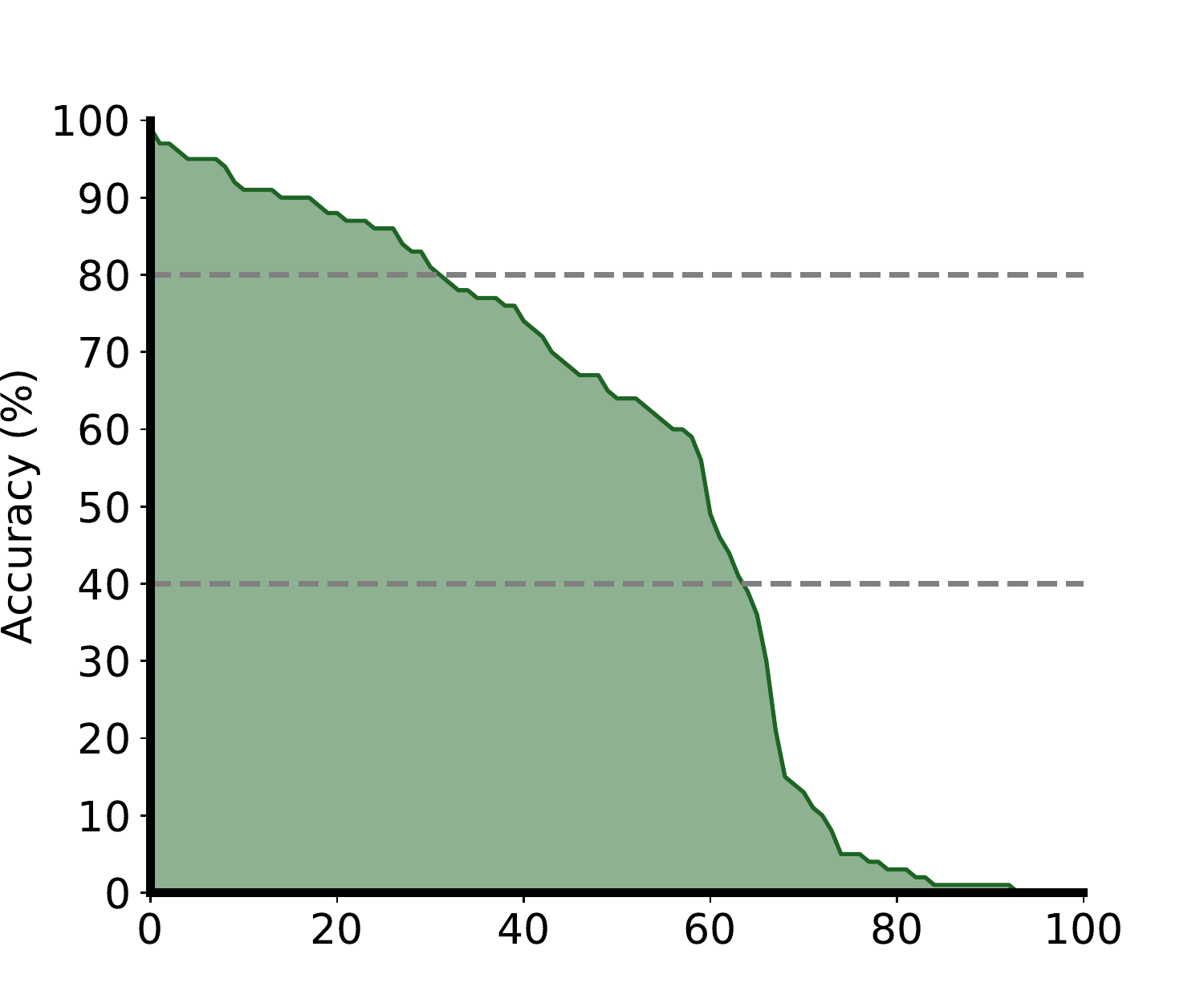}
    }
    \subfigure[DST w/o worst]{
        \includegraphics[width=0.3\textwidth]{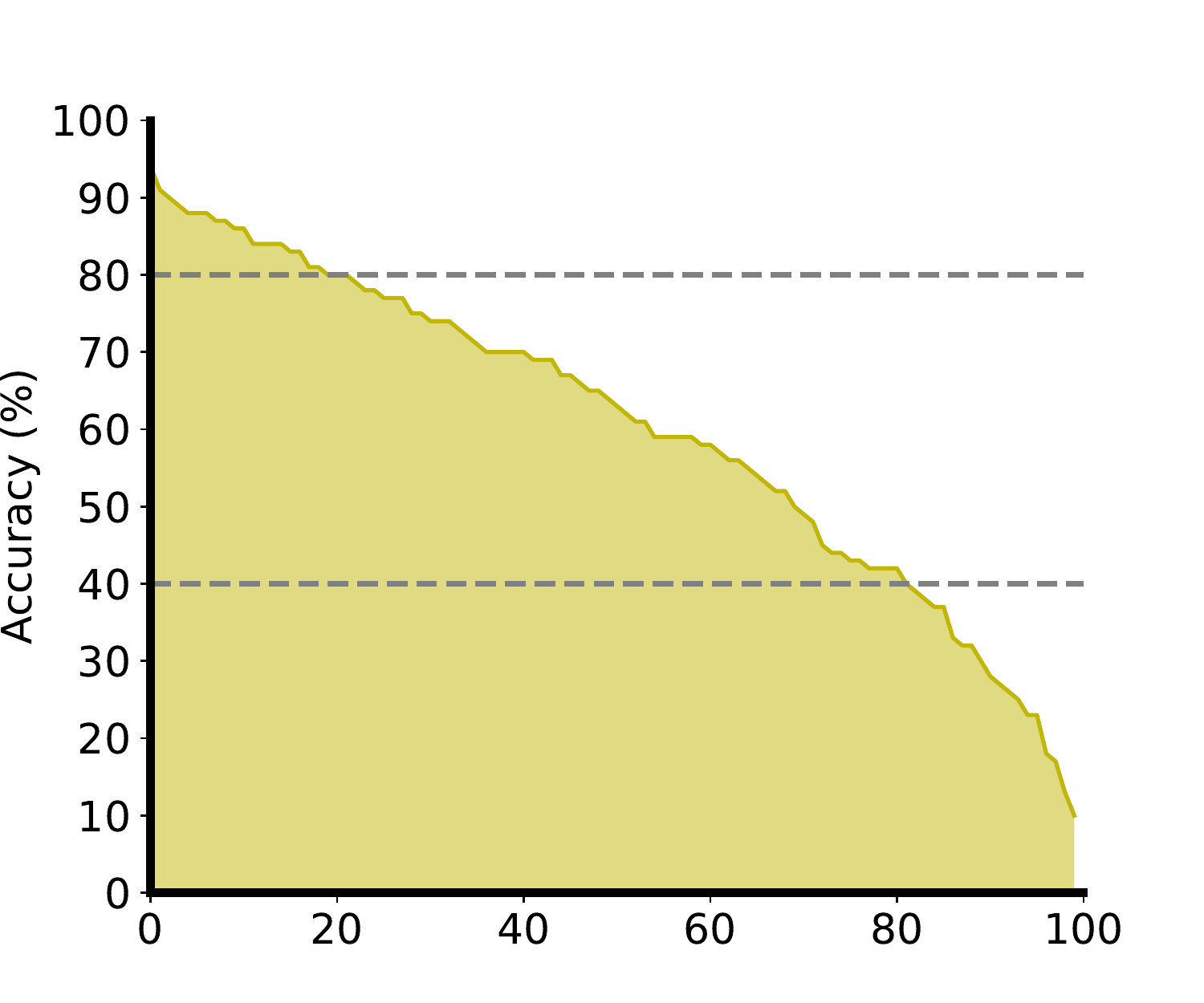}
    }
    \subfigure[DST]{
        \includegraphics[width=0.3\textwidth]{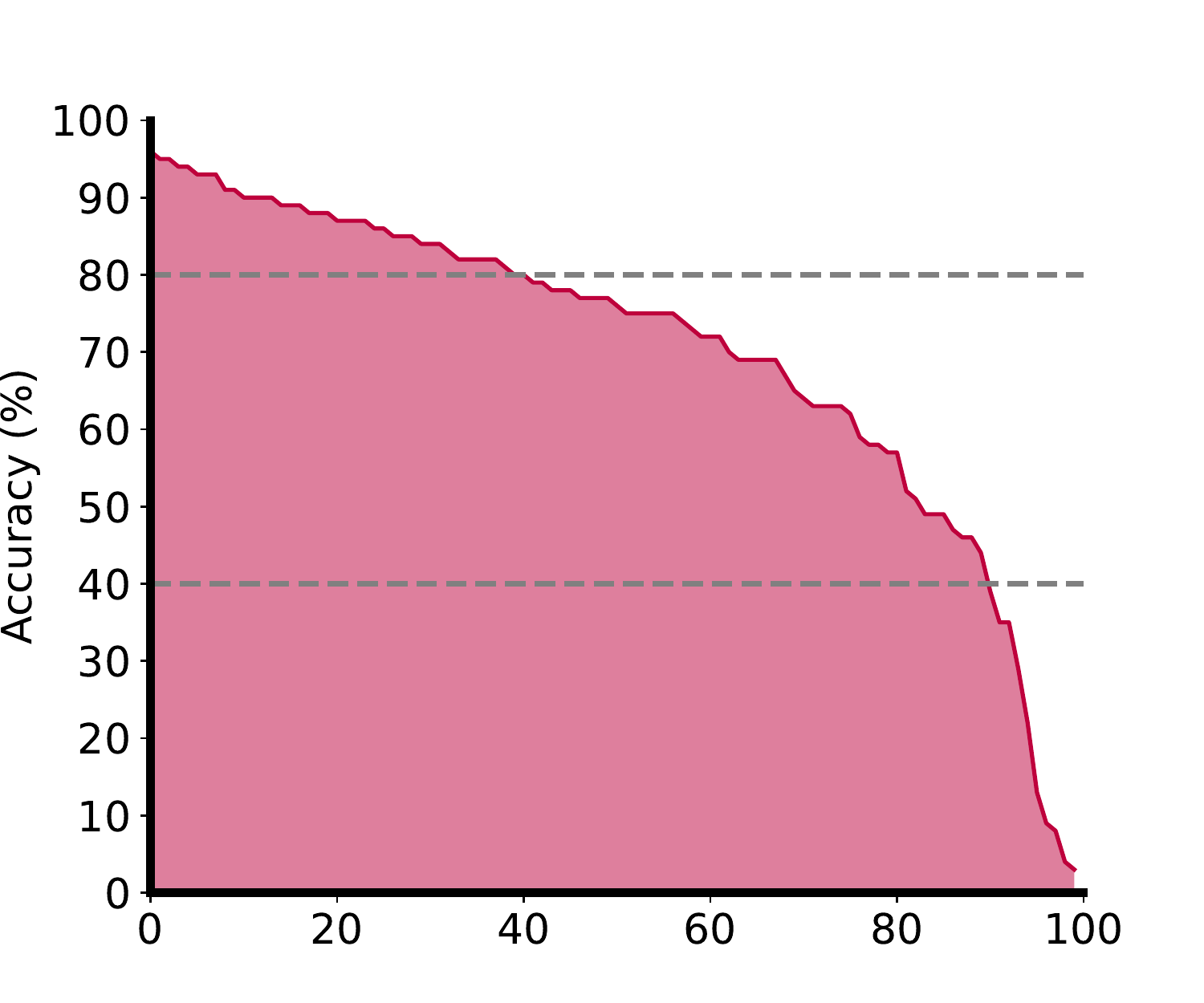}
    }
    \caption{Top-1 accuracy of each category on \emph{CIFAR-100} (ResNet50, supervised pre-trained). }
    \label{fig:appendix_per_cls}
\end{figure}

\begin{figure}[!ht]
    \centering
    \subfigure[FixMatch]{
        \includegraphics[width=0.3\textwidth]{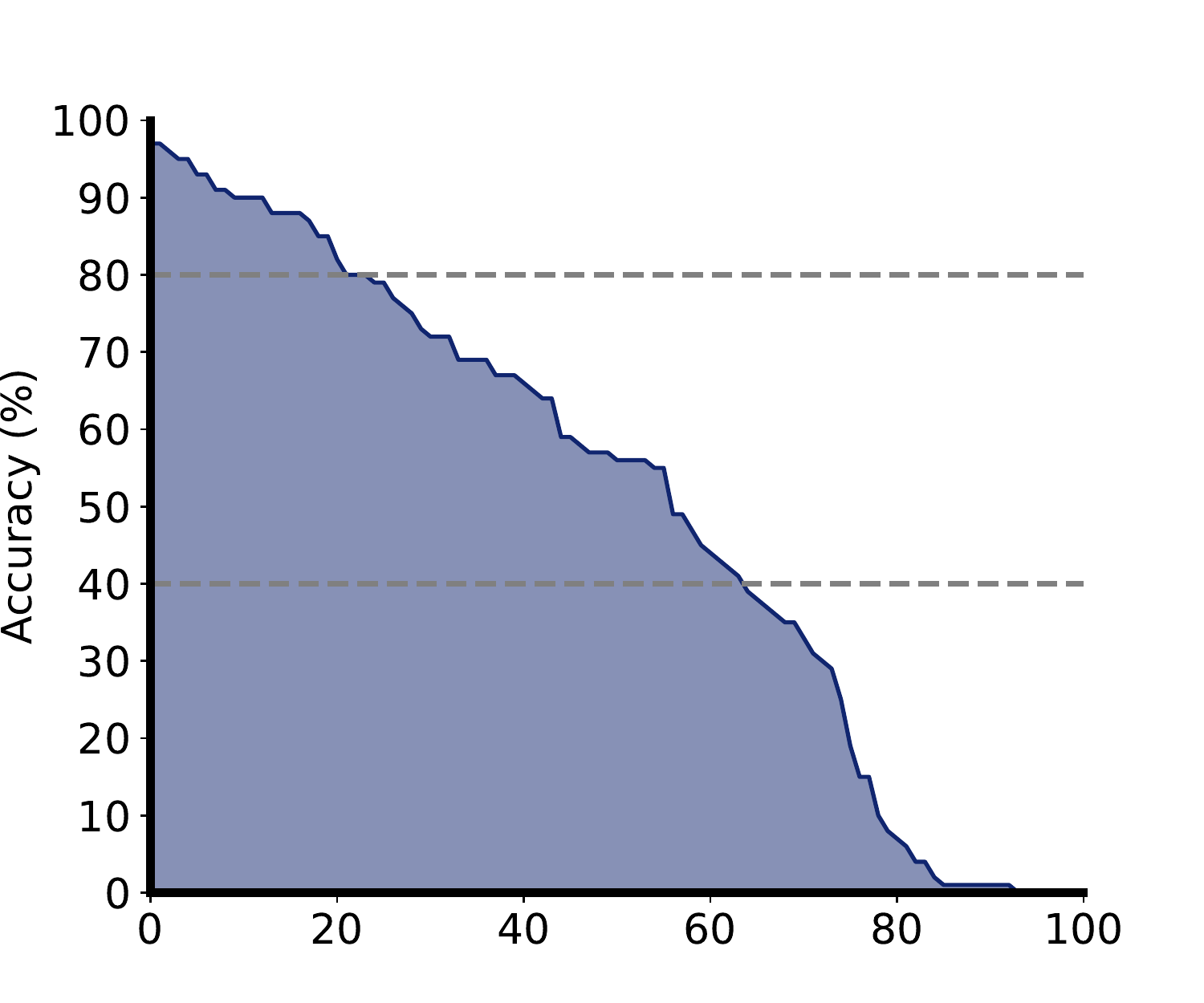}
    }
    \subfigure[DST]{
        \includegraphics[width=0.3\textwidth]{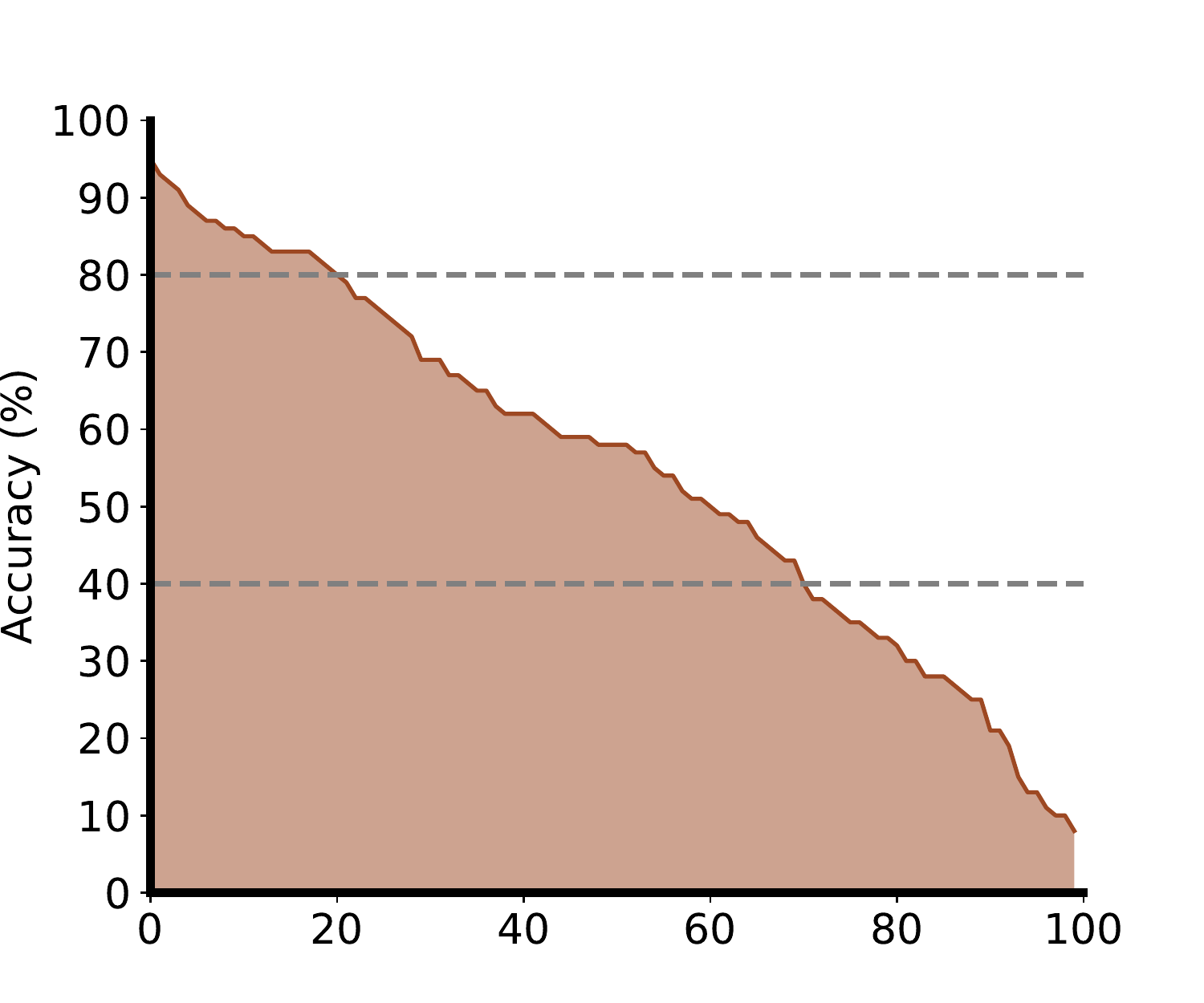}
    }
    \caption{Top-1 accuracy of each category in on \textit{CIFAR-100} (Wide ResNet-28-8, train from scratch). }
    \label{fig:scratch_acc_per_category}
\end{figure}

Figures \ref{fig:appendix_per_cls} and \ref{fig:scratch_acc_per_category} plot the top-1 accuracy of each category on \emph{CIFAR-100} yielded by self-training on $400$ labels and unlabeled data when using supervised pre-trained models or training from scratch, respectively. The results are consistent with our previous analysis (Section \ref{sec:experiments_analysis}) that {DST improves the performance of those poorly-behaved categories}.

\subsection{Experiments with varying amounts of labeled data}
Table \ref{table:more_supervision} reports the performance of DST with $1000$ labels on \emph{CIFAR-100} with different pre-trained models. DST outperforms FlexMatch and DebiasMatch under both settings.

\begin{table}[!ht]
\caption{Experiments with 10 labels per-class on \emph{CIFAR-100} (ResNet50). }
\label{table:more_supervision}
\addtolength{\tabcolsep}{-1pt}
\centering
\footnotesize
\begin{tabular}{@{}lcc@{}}
\toprule
             & Supervised Pre-Training & Unsupervised Pre-Training  \\ \midrule
Baseline     & 61.5                    & 56.2                 \\
Pseudo Label \cite{pseudo_label} & 67.4                    & 57.3                 \\
$\Pi$-Model \cite{Temporal_Ensembling}     & 63.3                    & 55.5                 \\
Mean Teacher \cite{Mean_Teacher} & 67.0                    & 63.5                 \\
UDA \cite{UDA}         & 65.1                    & 67.5                 \\
FixMatch \cite{FixMatch}     & 67.8                    & 64.2                 \\
Self-Tuning \cite{Self-Tuning}  & 66.0                    & 60.2                 \\
FlexMatch  \cite{FlexMatch}  & 71.2                    & 71.1                 \\
DebiasMatch \cite{DebiasMatch}  &  73.5                       &    73.9                 \\
\textbf{DST (FixMatch)}      & \textbf{75.6}           & \textbf{76.8}                 \\ \bottomrule
\end{tabular}
\end{table}

\begin{figure}[!htbp]
\centering
\includegraphics[width=0.35\textwidth]{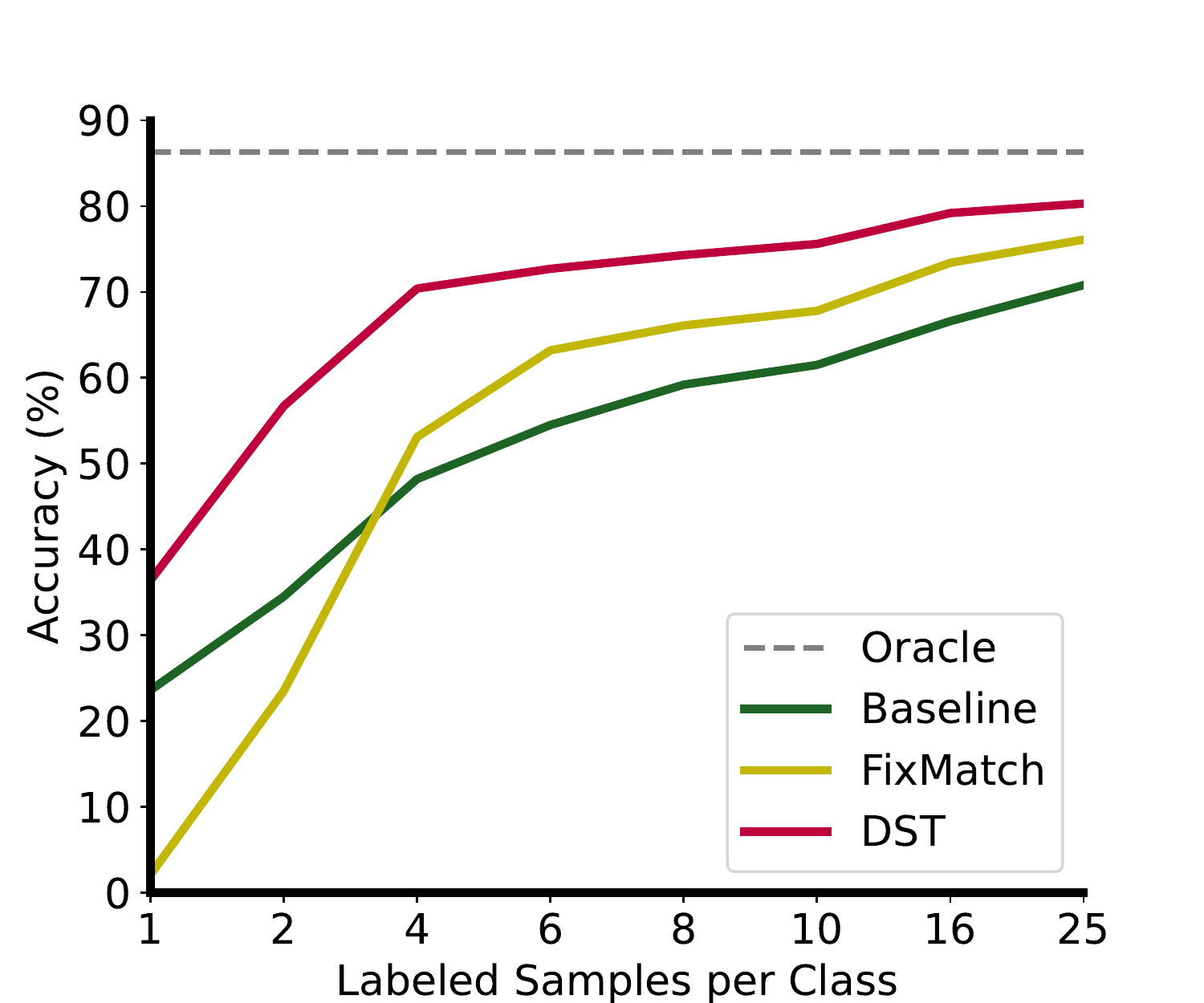}
\caption{Ablation on the amount of labeled data on \emph{CIFAR-100} (ResNet50, supervised pre-trained).}
\label{fig:ablation_label_size}
\end{figure}

{Figure \ref{fig:ablation_label_size} plots the accuracy of FixMatch and DST when the number of labeled samples per class varies from $1$ to $25$ on \emph{CIFAR-100} with supervised pre-trained models. Experiments show that {DST is less sensitive to the amount of labeled data} than FixMatch and yield consistent improvement.} 

\subsection{Analysis on the behavior of pseudo labels}
\label{appendix:analysis_pseudo}

\subsubsection{Using unsupervised pre-trained models}
\begin{figure}[!ht]
    \centering
    \vspace{-10pt}
    \subfigure[Quantity]{
        \includegraphics[width=0.3\textwidth]{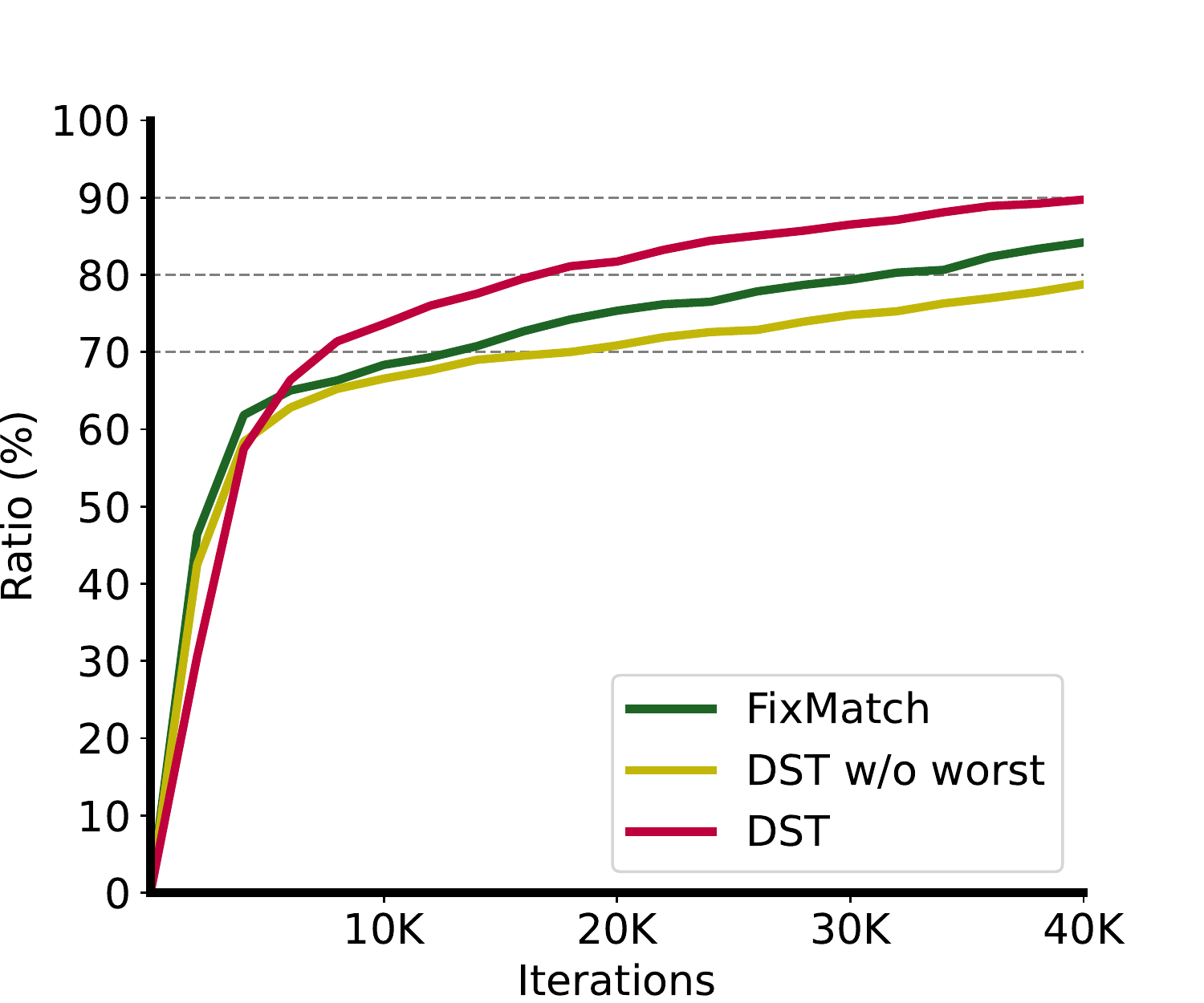}
    }
    \subfigure[Quality]{
        \includegraphics[width=0.3\textwidth]{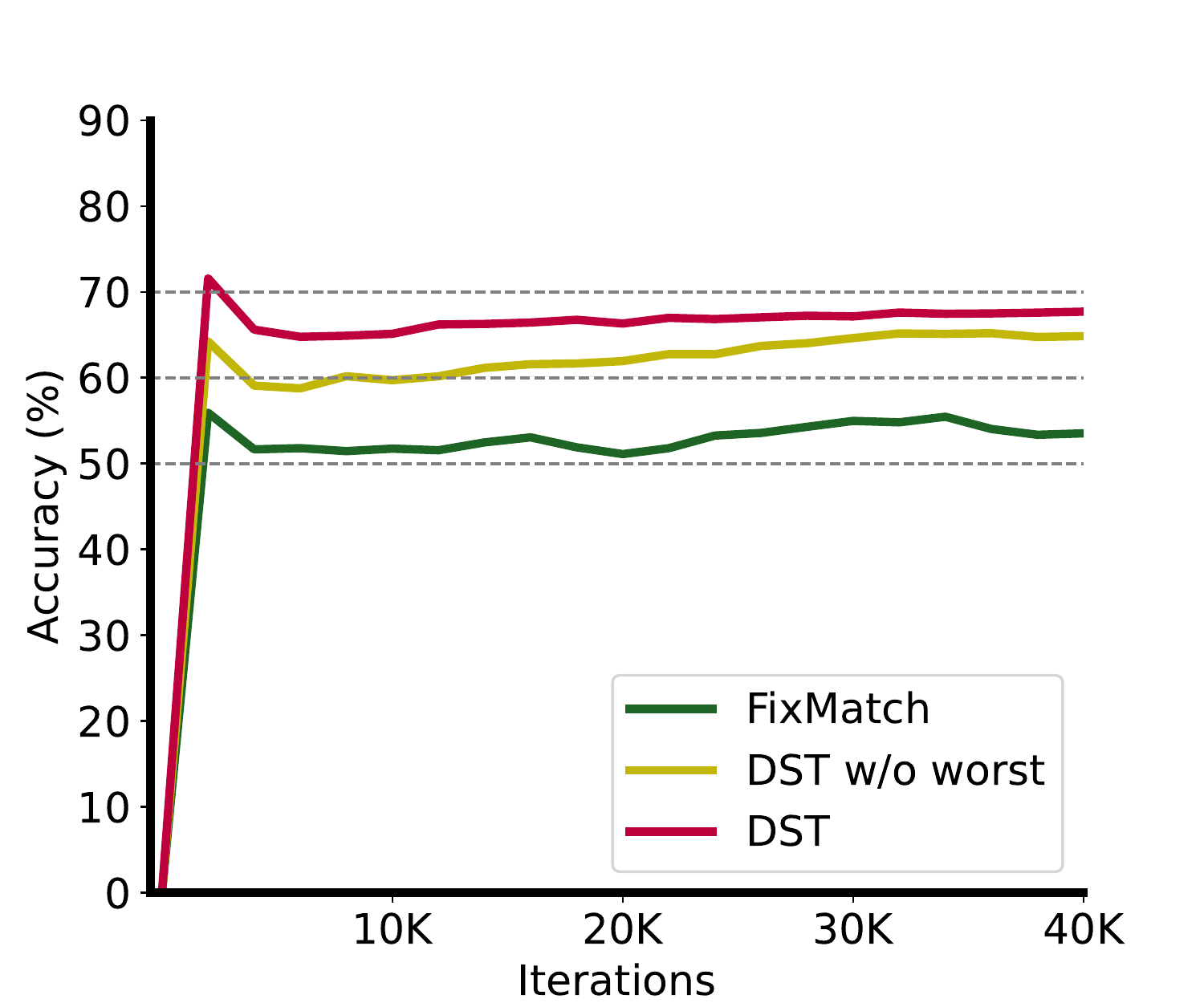}
    }
    \subfigure[Quantity of poorly-behaved categories]{
        \includegraphics[width=0.3\textwidth]{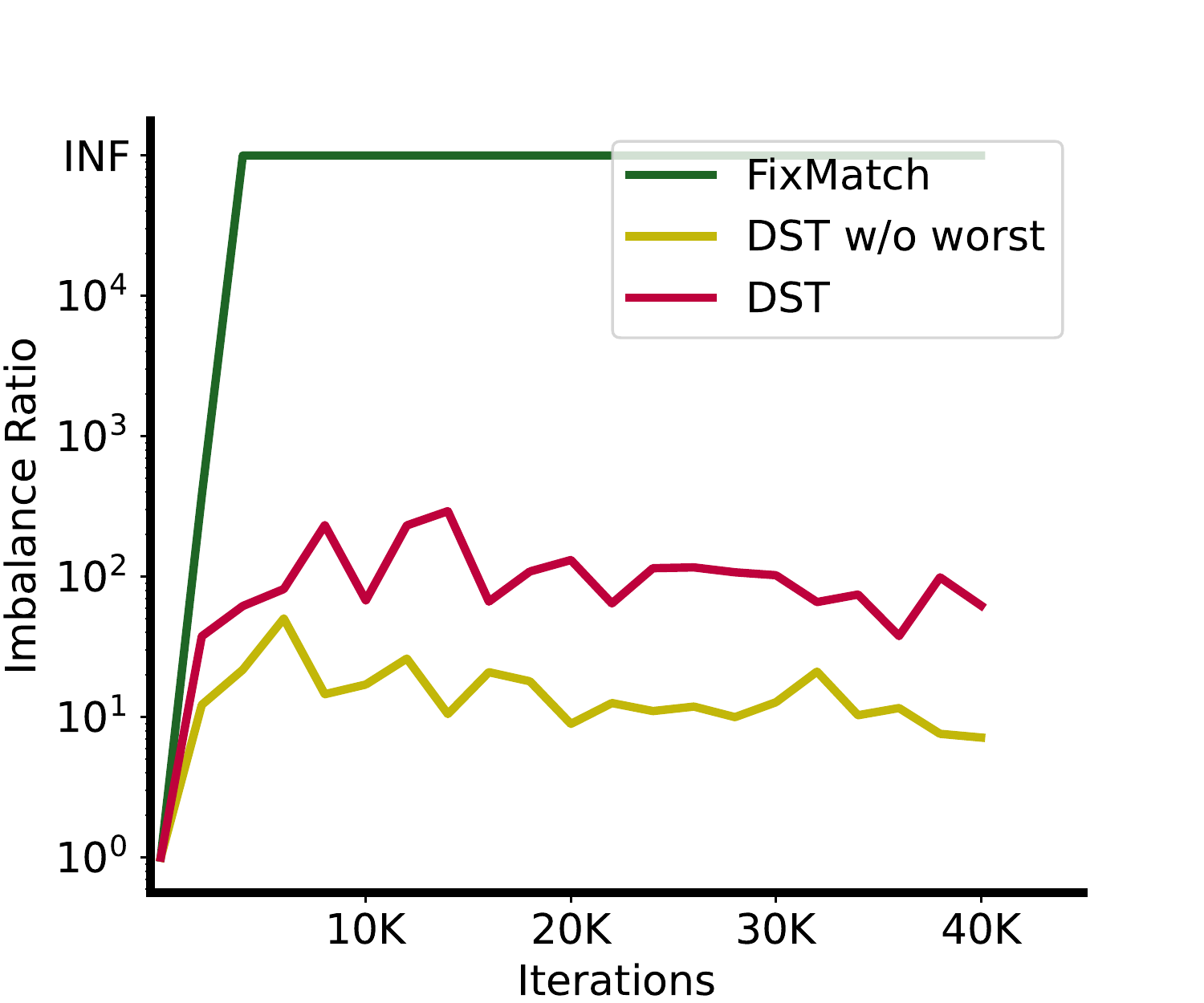}
    }
    \vspace{-5pt}
    \caption{Analysis on the behavior of pseudo labels on \emph{CIFAR-100} (ResNet50, unsupervised pre-trained). \textbf{(a)} The quantity of pseudo labels above the confidence threshold. \textbf{(b)} The accuracy of pseudo labels. \textbf{(c)} The class imbalance ratio $I$ of pseudo labels. } 
    \label{fig:appendix_moco}
    \vspace{-5pt}
\end{figure}

In this subsection, we explore the behavior of pseudo labels with unsupervised pre-trained models. Concretely, we focus on the quantity, accuracy as well as class imbalance ratio $I$ of pseudo labels. Recall that $I={{\max}_cN(c)}/{{\min}_{c'} N(c')}$,
where $N(c)$ denotes the number of predictions that fall into category $c$. Figure \ref{fig:appendix_moco} shows the results on \emph{CIFAR-100} with $400$ labels. We observe the same phenomenon that {DST effectively reduces the bias of pseudo labels, thereby improving the self-training process}.

\subsubsection{Comparison with other methods}

\begin{figure}[!hbtp]
\vspace{-10pt}
    \centering
    \subfigure[Comparison with methods that adopt dynamic threshold.]{
        \includegraphics[width=0.35\textwidth]{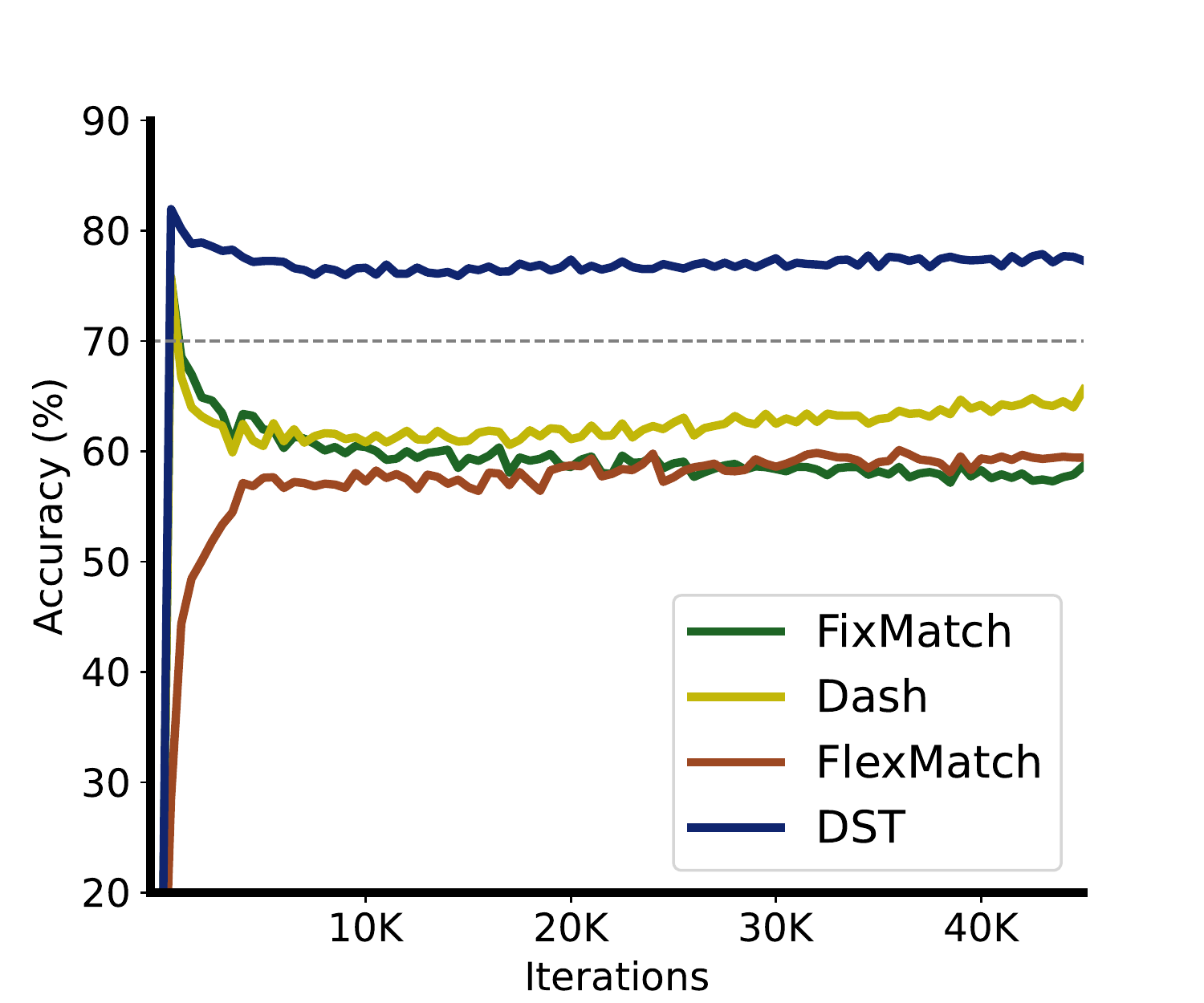}
    }
    \hspace{20pt}
    \subfigure[Comparison with methods that adopt multi-view training.]{
        \includegraphics[width=0.35\textwidth]{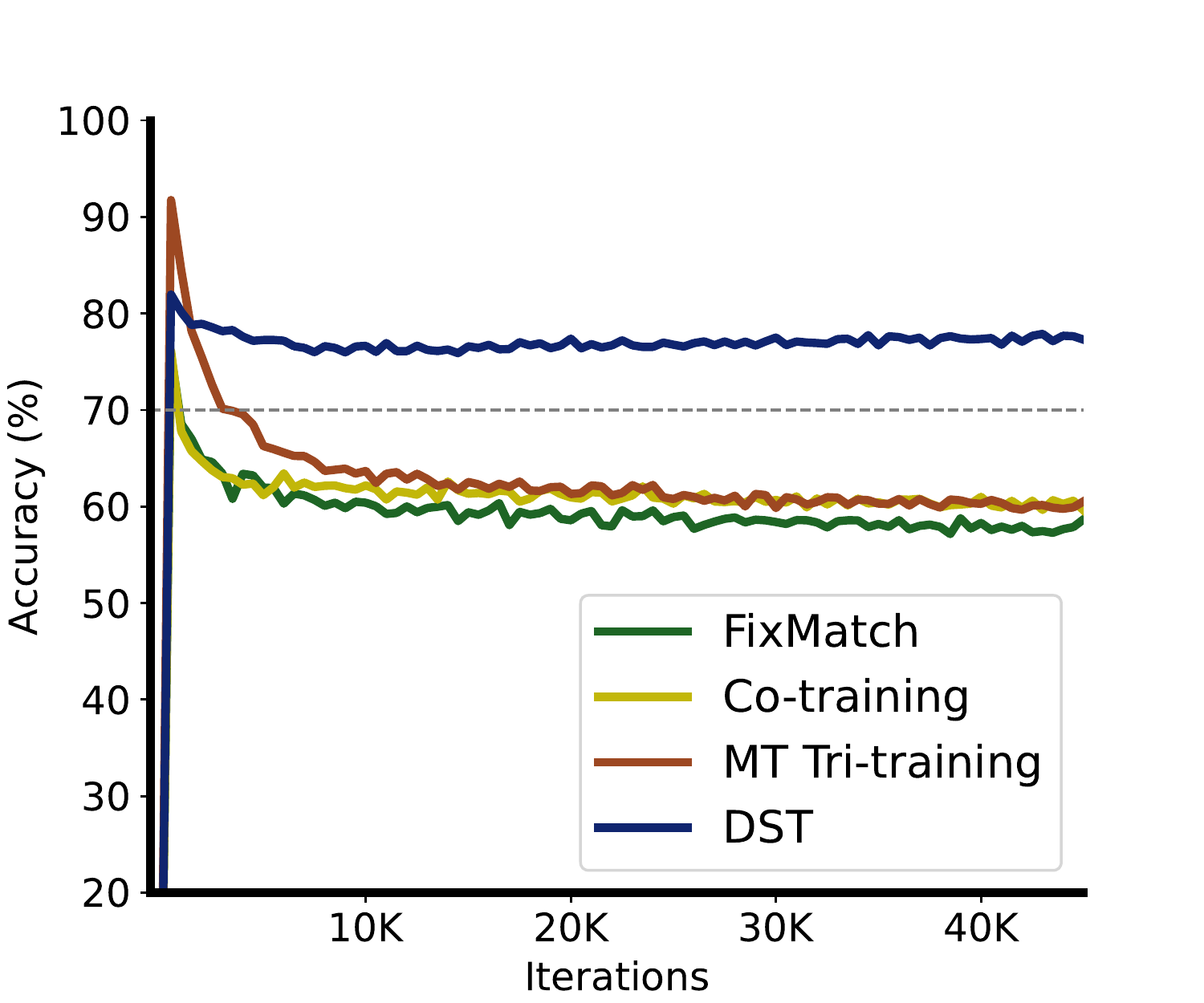}
    }
    \caption{Quality of pseudo labels (\emph{CIFAR-100}, ResNet50, supervised pre-trained).}
    \label{fig:comparison}
\vspace{-10pt}
\end{figure}

We consider two lines of work that promote the quality of pseudo labels by (1) using dynamic threshold, including Dash \cite{Dash} and FlexMatch \cite{FlexMatch}; (2) adopting multi-view training, including Co-training \cite{CoTraining} and Multi-task Tri-training \cite{MTTriTraining}. Table \ref{tab:comparison_improve_quality} shows that DST outperforms baselines by considerable margins. Figure \ref{fig:comparison} plots the quality of pseudo labels and reveals that our method can better debias pseudo labeling and improve the quality of pseudo labels.

\begin{table}[!htbp]
\addtolength{\tabcolsep}{-2pt}
\centering
\footnotesize
\caption{Comparison with other methods that improve the quality of pseudo labels (\emph{CIFAR-100}, ResNet50, supervised pre-trained).}
\label{tab:comparison_improve_quality}
\begin{tabular}{@{}cc|cc|cc|c@{}}
\toprule
\multicolumn{2}{l|}{} & \multicolumn{2}{c|}{Dynamic Thresholding} & \multicolumn{2}{c|}{Multi-head Training} & \multicolumn{1}{l}{} \\ \midrule
Baseline  & FixMatch  & Dash           & FlexMatch          & Co-training       & MT Tri-training      & DST             \\ \midrule
48.2      & 53.1      & 55.4           & 63.4               & 54.4              & 54.4                 & \textbf{70.4}                 \\ \bottomrule
\end{tabular}
\vspace{-10pt}
\end{table}

\subsection{Ablation study on nonlinear main classifier head}
Experiments suggest that using a nonlinear pseudo head improves performance. We further explore how things are going for the main head. As shown in Table \ref{appendix:nonlinear}, {using nonlinear main head or not results in a similar performance on average}. We conjecture this is because a nonlinear main head is more likely to over-fit with few labeled samples.

\begin{table}[!ht]
\caption{Ablation on nonlinear main head on \emph{CIFAR-100} (FixMatch, ResNet50, supervised pre-trained). }
\label{appendix:nonlinear}
\addtolength{\tabcolsep}{1pt}
\centering
\footnotesize
\begin{tabular}{l|cccc}
\toprule
\multirow{2}{*}{Head} & \multicolumn{2}{c}{Supervised Pre-Training} & \multicolumn{2}{c}{Unsupervised Pre-Training} \\
& 400 labels & 1000 labels & 400 labels & 1000 labels \\ \midrule
Linear & 53.1 & \textbf{67.8} & \textbf{51.4} & \textbf{64.2} \\
Nonlinear & \textbf{54.1} & 67.2 & 50.4 & 64.0 \\ 
\bottomrule
\end{tabular}
\end{table}

\section{Broader Impact}

First, our research helps improve the performance and training stability of various existing self-training methods, especially when the labeled data is scarce. 
Second, our proposed method is simple yet effective and thus can potentially reduce the labeling cost of many real-world machine learning applications. 
Finally, our research helps reduce the bias of self-training models
and improves their performance balance.

\end{document}